\documentclass[10pt,twocolumn,letterpaper]{article}

\usepackage{cvpr}
\usepackage{times}
\usepackage{epsfig}
\usepackage{graphicx}
\usepackage{amsmath, bm}
\usepackage{amssymb}
\usepackage{multirow}
\usepackage{microtype}
\usepackage[normalem]{ulem}
\useunder{\uline}{\ul}{}
\usepackage{color}
\usepackage{nicefrac}
\usepackage{comment}
\usepackage{enumitem}
\usepackage{adjustbox}
\usepackage{mathabx}
\usepackage{tabu}
\usepackage{booktabs}
\usepackage{diagbox}
\usepackage{cuted}
\usepackage{capt-of} 
\usepackage{multicol}
\usepackage[table,x11names]{xcolor}

\definecolor{gray}{RGB}{222,222,222}
\definecolor{nvgreen}{RGB}{118, 185, 0}

\newcommand{\midsepremove}{\aboverulesep = 0mm \belowrulesep = 0mm}
\midsepremove
\newcommand{\midsepdefault}{\aboverulesep = 0.605mm \belowrulesep = 0.984mm}
\midsepdefault

\usepackage{subcaption}
\usepackage{caption}
\newcommand{\sign}{\text{sign}}

\usepackage[pagebackref=false,breaklinks=true,letterpaper=true,bookmarks=false,colorlinks,citecolor=nvgreen,linkcolor=nvgreen]{hyperref}

\cvprfinalcopy 

\def\cvprPaperID{1591} 

\ifcvprfinal\fi

\begin{document}




\title{See through Gradients: Image Batch Recovery via GradInversion}

\author{
Hongxu Yin, Arun Mallya, Arash Vahdat, Jose M. Alvarez,
Jan Kautz, Pavlo Molchanov\\
\vspace{-0.35cm}
\\
NVIDIA \\
\tt\small \{dannyy, amallya, avahdat, josea, jkautz, pmolchanov\}@nvidia.com
}

\maketitle



\begin{abstract}
Training deep neural networks requires gradient estimation from 
data batches to update parameters. Gradients per parameter are averaged over a set of data and this has been presumed to be safe for privacy-preserving training in joint, collaborative, and federated learning applications. 
Prior work only showed the possibility of recovering input data given gradients under very restrictive conditions --  a single input point, or a network with no non-linearities, or a small $32\times32$ px input batch. Therefore, averaging gradients over larger batches was thought to be safe. In this work, we introduce GradInversion, using which input images from a larger batch 
($8$ -- $48$ images) 
can also be recovered for large networks such as ResNets ($50$ layers), on complex datasets such as ImageNet ($1000$ classes, $224\times224$ px). We formulate an optimization task that converts random noise into natural images, matching gradients while regularizing image fidelity. 
We also propose an algorithm for target class label recovery given gradients. We further propose a group consistency regularization framework, where multiple agents starting from different random seeds work together to find an 
enhanced reconstruction of the original data batch. 
We show that gradients encode a surprisingly large amount of information, such that all the individual images can be recovered with high fidelity via GradInversion, even for complex datasets, deep networks, and large batch sizes.

\vspace{-5mm}
\end{abstract}


\section{Introduction}

\begin{figure}[t]
\centering
\renewcommand*{\arraystretch}{0.3}
\begin{tabular}{@{}c@{}}
\includegraphics[width=\linewidth]{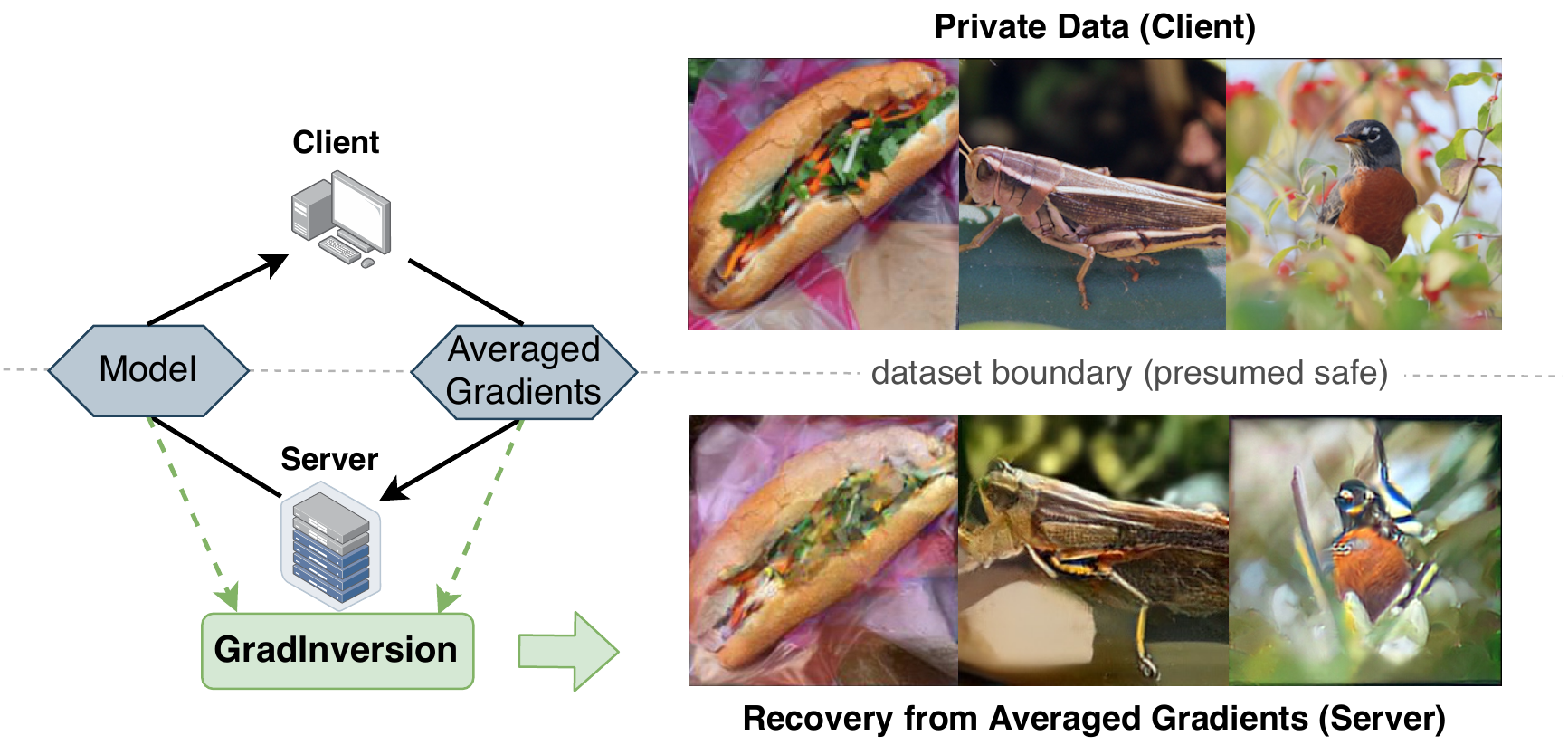} \\ \vspace{0.5mm}
\small{(a) Inverting averaged gradients to recover original image batches} \vspace{0.6mm} \\
\includegraphics[width=1.0\linewidth,trim={.7cm 0.0cm 0 0}, clip]{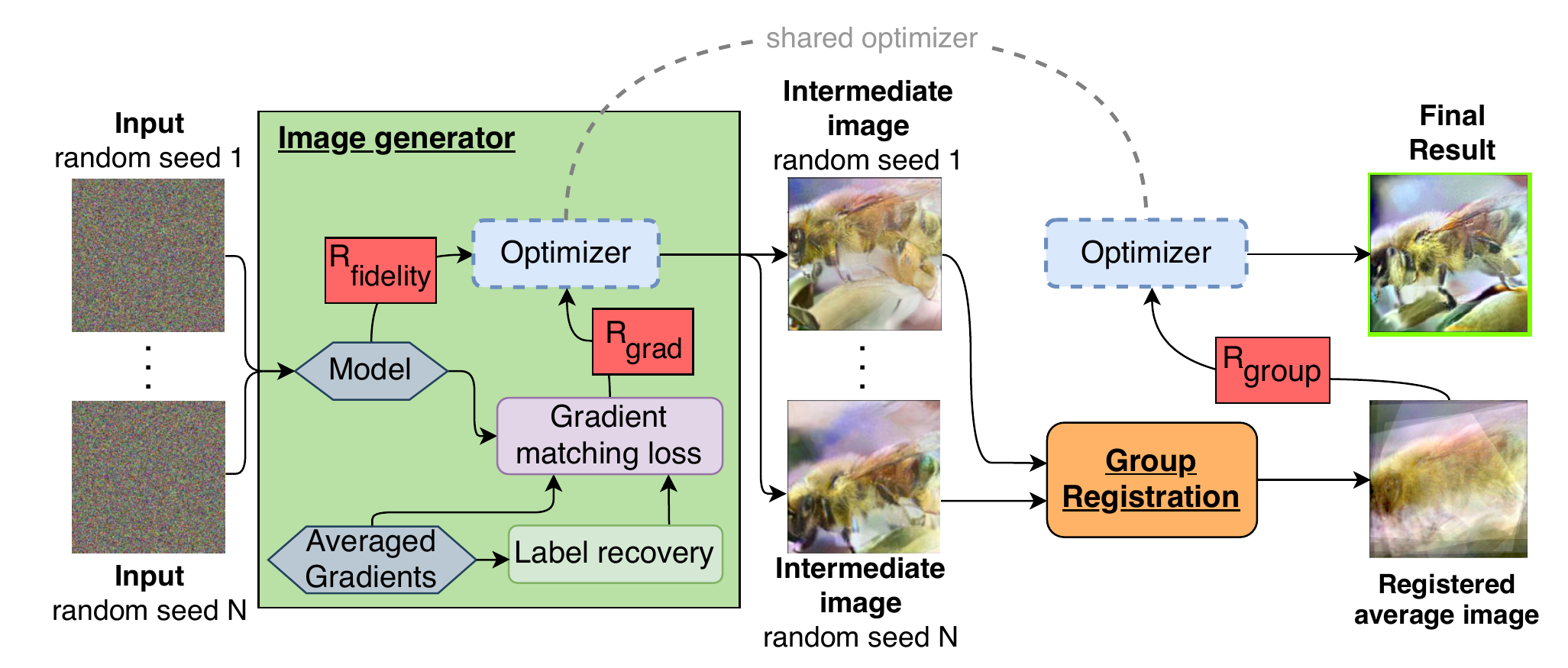} \\
\small{(b) Overview of our proposed GradInversion method} \vspace{1mm} \\\\
\end{tabular}
\caption{We propose (a) GradInversion to recover hidden training image batches with high fidelity via inverting \textit{averaged} gradients. GradInversion formulates (b) an optimization process that transforms noise to input images (Sec.~\ref{sec:obj_function}). It starts with label restoration from the gradient of the fully connected layer (Sec.~\ref{sec:label_restore}), then optimizes inputs to match target gradients under fidelity regularization (Sec.~\ref{sec:r_feature}) and registration-based group consistency regularization (Sec.~\ref{sec:r_group}) to improve reconstruction quality. This enables recovery of $224\times 224$ pixel ImageNet samples from ResNet-50 batch gradients, which was previously impossible (please zoom into examples above. More in Sec.~\ref{sec:exp}).
} 
\label{fig:teaser}
\vspace{-1mm}
\end{figure}

Sharing weight updates or gradients during training
is the central idea behind collaborative, distributed, and federated learning of deep networks~\cite{bonawitz2019towards, iandola2016firecaffe,konevcny2016federated, konevcny2016federated2, li2014scaling}. 
In the basic setting of federated stochastic gradient descent, each device learns on local data, and shares gradients to update a global model.
Alleviating the need to transmit training data offers several key advantages.
This keeps user data private, allaying concerns related to user privacy, security, and other proprietary concerns. 
Further, this eliminates the need to store, transfer, and manage possibly large datasets.
With this framework, one can train a  model on medical data without access to any individual's data~\cite{brisimi2018federated, mcmahan2017communication}, or perception model for autonomous driving without invasive data collection~\cite{samarakoon2018federated}.

While this setting might seem safe at first glance, a few recent works have begun to question the central premise of federated learning - is it possible for gradients to leak private information of the training data?
Effectively serving as a  ``proxy'' of the training data, the link between gradients to the data in fact offers potential for retrieving information: from revealing the positional distribution of original data~\cite{melis2019exploiting,shokri2017membership}, to even enabling pixel-level detailed image reconstruction from gradients~\cite{ geiping2020inverting, zhao2020idlg, zhu2019deep}. Despite remarkable progress, inverting an original image through gradient matching remains a very challenging task -- successful reconstruction of images of high resolution for complex datasets such as ImageNet~\cite{deng2009imagenet} has remained elusive for batch sizes larger than one. 

Emerging research on network inversion techniques offers insights into this task. Network inversion enables noise-to-image conversion via back-propagating gradients on appropriate loss functions to the learnable inputs. 
Initial solutions were limited to shallow networks and low-resolution synthesis~\cite{fredrikson2015modelinversionattack, nguyen2016synthesizing}, or creating an artistic effect~\cite{mordvintsev2015deepdream}. However, the field has rapidly evolved, enabling high-fidelity, high-resolution image synthesis on ImageNet from commonly trained classifiers, making downstream tasks data-free for pruning, quantization, continual learning, knowledge transfer, \etc~\cite{cai2020zeroq, haroush2020knowledge,santurkar2019image,yin2020dreaming}. Among these, DeepInversion~\cite{yin2020dreaming} yields state-of-the-art results on image synthesis for ImageNet. It enables 
the synthesis of realistic data
from a vanilla pretrained ResNet-50~\cite{he2016deep} classifier by regularizing feature distributions through batch normalization (BN) priors. 

Building upon DeepInversion~\cite{yin2020dreaming}, we delve into the problem of batch recovery via gradient inversion. We formulate the task as the optimization of the input data such that the gradients on that data match the ones provided by the client, while ensuring realism of the input data. However, since the gradient is also a function of the ground-truth label, one of the main challenges is to identify the ground-truth label for each data point in the batch. To tackle this, we propose a one-shot batch label restoration algorithm that uses gradients from the last fully connected layer. 

Our goal is to recover the exact images that the client possesses. By starting from noisy inputs generated by different random seeds, multiple optimization processes
are likely to converge to different minimas. Due to the inherently spatially-invariant nature of convolutional neural networks (CNNs), these resulting images share spatial information but differ in the exact location and arrangement.
To allow for improved convergence towards the ground truth images, we compute a registered mean image from all candidates and introduce a group consistency regularization term on every optimization process to reduce deviation. 
We find that the proposed approach and group consistency regularization provide superior better image recovery compared to prior optimization approaches~\cite{geiping2020inverting, zhu2019deep}.

Our non-learning based image recovery method recovers more specific details of the hidden input data when compared to 
the state-of-the-art generative adversarial networks (GAN), such as BigGAN~\cite{brock2018biggan}. More importantly, we demonstrate that a full recovery of individual images of $224\times224$ px resolution with high fidelity and visual details, by inverting gradients of the batch, is now made feasible even up to batch size of $48$ images. 


Our main contributions are as follows:
\begin{itemize}[topsep=1pt,itemsep=1pt,partopsep=0pt, parsep=0pt,leftmargin=\labelwidth]

    \item We introduce GradInversion to recover hidden original images from random noise via optimization given batch-averaged gradients.
    
    \item We propose a label restoration method to recover ground truth labels 
    using final fully connected layer gradients.

    \item We introduce a group consistency regularization term, based on multi-seed optimization and image registration, to improve reconstruction quality.
    
    
    \item We demonstrate that a full recovery of detailed individual images from batch-averaged gradients is now feasible for deep networks such as ResNet-50.
    
    \item We introduce a new \textit{Image Identifiability Precision} metric to measure the ease of inversion over varying batch sizes, and identify samples vulnerable to 
    inversion.
    
\end{itemize}
\section{Related Work}

\noindent\textbf{Image synthesis.} 
GANs~\cite{gulrajani2017improved, karras2020analyzing,miyato2018spectral,nguyen2017plug,zhang2018self} have delivered state-of-the-art results for generative image modeling, \eg, BigGAN-deep on ImageNet~\cite{brock2018biggan}. Training a GAN's generator, however, requires access to original data. Multiple works have also looked into training GANs given only a pretrained model~\cite{chen2019data, micaelli2019zero}, but result in images that lack details or perceptual similarities to original data. 

Prior work in security studies image synthesis from a pretrained single network. The \textit{model inversion} attack by Fredrikson \etal \cite{fredrikson2015modelinversionattack} optimizes inputs to obtain class images using gradients from the target model. Follow-up works~\cite{he2019model, wang2015regression,yang2019adversarial} scale to new threat scenarios, but remain limited to shallow networks. The \textit{Secret Revealer}~\cite{zhang2020secret} exploits priors from auxiliary datasets and trains GANs to guide inversion, scales the attack to modern architectures, but on the datasets
with less diverse samples, \eg, MNIST and face recognition.

Though originally aiming at understanding network properties, visualization techniques offer another viable option to generate images from networks. Mahendran \etal \cite{mahendran2016visualizing} explore inversion, activation maximization, and caricaturization to synthesize ``natural pre-images'' from a trained network
~\cite{mahendran2015understanding,mahendran2016visualizing}. Nguyen \etal use global generative priors to help invert trained networks~\cite{nguyen2016synthesizing} for images, followed by Plug \& Play~\cite{nguyen2017plug} that boosts up image diversity and quality via latent priors. 
These methods still rely on auxiliary dataset information, feature embedding, or altered training.

Recent efforts focus on image generation from a pretrained network without any auxiliary information. DeepDream~\cite{mordvintsev2015deepdream} by Mordvintsev \etal hints on ``dreaming'' new visual features onto images leveraging gradients on inputs, extendable towards noise-to-image conversion. Saturkar \etal~\cite{santurkar2019image} extended the approach to more realistic images.
The more recent extensions~\cite{cai2020zeroq,yin2020dreaming} significantly improved state-of-the-art performance on image synthesis from off-the-shelf classifiers, without auxiliary information nor additional training but relying on BN statistics. 

\noindent\textbf{Gradient-based inversion.} 
There have been early attempts to invert gradients in pursuit of proxy information of the  original data, \eg, the existence of certain training samples~\cite{melis2019exploiting,shokri2017membership} or sample properties~\cite{hitaj2017deep,shokri2017membership} of the dataset. These methods primarily focus on very shallow networks.

A more challenging task aims at reconstructing the exact images from gradients. The early attempt by Phong \etal~\cite{le2017privacy} brought theoretical insights on this task by showing provable reconstruction feasibility on single neuron or single layer networks. Wang \etal~\cite{wang2019beyond} empirically inverted out single image representations from gradients of a $4$-layer network. Along the same line, Zhu \etal~\cite{zhu2019deep} pushed gradient inversion towards deeper architectures by jointly optimizing for ``pseudo'' labels and inputs to match target gradients. The method leads to accurate reconstruction up to pixel-level, while remains limited to continuous models (\eg, ones with sigmoid instead of ReLU) without any strides, and scales up to 
low-resolution CIFAR datasets. Zhao \etal~\cite{zhao2020idlg} extend the approach with a label restoration step, hence improving speed of single image reconstruction. The very recent work by Geiping \etal~\cite{geiping2020inverting} for the first time pushed the boundary towards ImageNet-level gradient inversion - it reconstructs single images from gradients. Despite remarkable progress, the field struggles on ImageNet for any batch size larger than one, when gradients get averaged. 

\section{GradInversion}
In this section, we explain GradInversion in detail.
We first frame the problem of input reconstruction from gradients as an optimization process. Then, we explain our batch label restoration method, followed by the auxiliary losses used to ensure realism and group consistency regularization. 

\subsection{Objective Function}
\label{sec:obj_function}
Given a network with weights $\mathbf{W}$ and a batch-averaged gradient $\Delta \mathbf{W}$ 
calculated from a ground truth 
batch with images $\mathbf{x^*}$ and labels $\mathbf{y^*}$, our optimization solves for
\begin{equation}
\mathbf{\mathbf{\hat{x}^*}} = \underset{\mathbf{\hat{x}}}{\text{argmin }}{ {\mathcal{L_\text{grad}}(\mathbf{\hat{x}}; \mathbf{W}, \Delta \mathbf{W}) +  \mathcal{R_\text{aux}}(\mathbf{\hat{x})}}}, 
\label{eqn:main_error}
\end{equation}
\noindent
where $\mathbf{\hat{x}}\in \mathbb{R}^{K\times C \times H\times W}$ ($K, C, H, W$ being the batch size, number of color channels, height, width) is a ``synthetic'' input batch, initialized as random noise and optimized towards the ground truth $\mathbf{x^*}$. $\mathcal{L_\text{grad}}(\cdot)$ enforces matching of the gradients of this synthetic data (on the original loss for a network with weights $\mathbf{W}$) with the provided gradients $\Delta \mathbf{W}$.
This is augmented by auxiliary regularization $\mathcal{R_\text{aux}}(\cdot)$ based on image fidelity and group consistency regularization,
\begin{equation}
\mathcal{R_\text{aux}}(\mathbf{\hat{x}) = \mathcal{R_\text{fidelity}}(\mathbf{\hat{x}}) + \mathcal{R_\text{group}}(\mathbf{\hat{x}})}. 
\label{eqn:r_aux}
\end{equation}
\noindent
\noindent

We next elaborate on each term individually. For gradient matching, we minimize the $\ell_2$ distances between gradients on the synthesized images $\mathbf{\hat{x}}$ and the ground truth gradient:
 \begin{equation}
\resizebox{.9\columnwidth}{!}{$\mathcal{L_\text{grad}}(\mathbf{\hat{x}}; \mathbf{W}, \Delta \mathbf{W}) = \alpha_{\text{G}} \sum_{l} || \nabla_{\mathbf{W}^{(l)}}\mathcal{L}(\mathbf{\hat{x}}, \mathbf{\hat{y}}) - \Delta \mathbf{W}^{(l)} ||_2$,}
\label{eqn:lgrad}
\end{equation}
\noindent
where $\Delta \mathbf{W}^{(l)} = \nabla_{\mathbf{W}^{(l)}}\mathcal{L}(\mathbf{x^*}, \mathbf{y^*})$ refers to ground truth gradient at layer $l$, and the summation, scaled by $\alpha_{\text{G}}$, runs over all layers. One key yet missing component here is the $\mathbf{\hat{y}}$ that initiates the backpropagation. We next explain an effective algorithm for restoring batch-wise label from the gradients of the fully connected classification layer.

\subsection{Batch Label Restoration} 
\label{sec:label_restore}
Considering the cross-entropy loss for the classification task, the ground truth gradient of $\mathbf{x}^* = [x_1, x_2, ..., x_K]$ of batch size $K$ can be decomposed into:
\begin{equation}
    \nabla_{\mathbf{W}}\mathcal{L}(\mathbf{x}^*, \mathbf{y}^*) = \frac{1}{K}\sum_{k}\nabla_{\mathbf{W}}\mathcal{L}(x_k, y_k),
    \label{eqn:lrestore_main}
\end{equation}
where ${x_k, y_k}$ denotes an original image/label pair. For each image $x_k$, the gradient \wrt the network final logits $z$ at index $n$ is $\nabla_{z_{n,k}}\mathcal{L}(x_k, y_k) = p_{k,n} - y_{k,n}$, where $p_{k,n}$ is the post-softmax probability in range ($0$, $1$), and $y_{k,n}$ is the binary presentation of $y_{k}$ at index $n$ among $N$ total classes. 
Consequentially, this leaves $\sign\big(\nabla_{z_{n,k}}\mathcal{L}(x_k, y_k)\big)$ negative \textit{iff} $n = n_{k}^*$ at the ground truth index, and positive otherwise. However, we do not have access to $\nabla_{z_{n,k}}\mathcal{L}(x_k, y_k)$ as gradients are only given \wrt the model parameters. 

Denote the parameters of the final fully connected classification layer by 
$\textbf{W}^{\text{(FC)}} \in \mathbb{R}^{M \times N}$ with $M$ being the number of embedded features, and $N$ being the number of target classes. Define $\Delta \mathbf{W}^{\text{(FC)}}_{m, n, k} := \nabla_{w_{m,n}}\mathcal{L}(x_k, y_k)$ as the gradient of the training loss for image $x_k$ \wrt $\textbf{W}_{m, n}^{\text{(FC)}}$, connecting feature $m$ to logits $n$. We are only given the average of the tensor $\Delta \mathbf{W}^{\text{(FC)}}$ along the batch dimension $k$. Using the chain rule we have:
\begin{equation}
\begin{aligned}
\Delta \mathbf{W}^{\text{(FC)}}_{m, n, k} = \nabla_{z_{n,k}}\mathcal{L}(x_k, y_k) \times \frac{\partial z_{n,k}}{\partial w_{m,n}}.
\end{aligned}
\end{equation}
Note that $\frac{\partial z_{n,k}}{\partial w_{m,n}} = o_{m,k}$ where $o_{m,k}$ is the  $m^{\text{th}}$ input of the fully connected layer, and is also the $m^{\text{th}}$ output of previous layer. If the previous layer has commonly used activation functions such as ReLU or sigmoid, $o_{m,k}$ is always non-negative. This hints on target label existence via signs of a new informative indicator: 
\begin{equation}
\small
\begin{aligned}
  \mathbf{S}_{n,k} := \sum_m \Delta \mathbf{W}^{\text{(FC)}}_{m, n, k} = \sum_{m}\underbrace{\nabla_{z_{n,k}}\mathcal{L}(x_k, y_k)}_\textrm{neg. \textit{iff} $n$ = $n_k^*$} \times \underbrace{o_{m,k}}_\textrm{non-neg.},
 \end{aligned}
\label{eqn:s_vec}
\end{equation}
\noindent
where $\mathbf{S} =\{\mathbf{S}_{n,k}\}$ is an $N \times K$-matrix, constructed by summing the tensor $\Delta \mathbf{W}^{\text{(FC)}}$ along the feature dimension $m$. Interestingly, $\mathbf{S}$ contains negative values for the ground truth label of each instance. Thus, the $k^{th}$ column of $\mathbf{S}$ can be used to restore the ground truth label for the $k^{th}$ image by simply identifying the index of the negative entry.
Zhao \etal~\cite{zhao2020idlg} explored this rule for single image label restoration. However, we do not have access to $\mathbf{S}$ in our multi-sample batch setup as the given gradients are averaged over all images.

Motivated by this, we define the $N$-dimensional batch-level vector $\mathbf{s} = \{ \mathbf{s}_n \}$ by averaging $\mathbf{S}$ along its columns:
\begin{equation}
\begin{aligned}
  \mathbf{s}_{n} := \frac{1}{K} \sum_{k} \mathbf{S}_{n,k} = \sum_{m} \big( \underbrace{
    \frac{1}{K} \sum_{k} \Delta \mathbf{W}^{\text{(FC)}}_{m, n, k}}_{\textrm{given in } \Delta \mathbf{W}^{\text{(FC)}}} \big).
\label{eqn:sbatch}
\end{aligned}
\end{equation}
\noindent
The appealing property of $\mathbf{s}$ is that it can be computed easily from the given gradient for the fully connected layer by summing it along the feature dimension $m$ as shown on the right hand side of Eqn.~\ref{eqn:sbatch}.

As noted above, each column in $\mathbf{S}$ is a vector, containing a single negative peak at the label index and positive otherwise. Since the vector $\mathbf{s}$ is a linear super-position of $\mathbf{S}$'s columns, from all individual images $x_k$'s in the batch, this information can be lost during summation. However, we empirically observe that the encoded positions often possess larger magnitudes $ |\mathbf{S}_{n_k^*, k}| \gg  |\mathbf{S}_{n\neq n_k^*, k}|$. 
This leaves a negative sign mostly intact when the summation brings in positive values from other images. 

To further enable a more robust propagation of negative signs, we utilize column-wise minimum values, instead of summation along the feature dimension for the $\mathbf{s}$ calculation: to have a sum along the feature dimension be negative, at least one of its positions has to be negative, but not vice versa. This further boosts up the label restoration accuracy, especially when the batch size is large. Thus, we formulate the final label restoration algorithm for batch size $K$ as:
%
\begin{equation}
\begin{aligned}
  \mathbf{\hat{y}} =  \text{arg sort}\big(\underset{m}{\text{min }}  \nabla_{\mathbf{W}_{m, n}^{\text{(FC)}}}\mathcal{L}(\mathbf{x^*}, \mathbf{y^*})\big)[:K],
\end{aligned}
\label{eqn:label_restore}
\vspace{-1mm}
\end{equation}
\noindent
with $m$ corresponding to the feature embedding dimension before the fully connected layer.
The resulting $\mathbf{\hat{y}}$ supports Eqn.~\ref{eqn:lgrad} in subsequent $\mathbf{\hat{x}}$ optimization in pursuit for $\mathbf{x^*}$. One limitation of the proposed method is that it assumes non-repeating labels in the batch, which generally holds for a randomly sampled batch of size $K$ that is much smaller than the number of classes at $1000$ for ImageNet.

Even with correct $\mathbf{y}^*$, finding the global minima for $\mathcal{L_\text{grad}}(\cdot)$ remains challenging. The task is under-constrained, 
suffers from information loss due to non-linearity and pooling layers, and has only one 
correct 
solution~\cite{geiping2020inverting,zhu2019deep}. We next introduce $\mathcal{R_\text{aux}}(\cdot)$ based on fidelity and group consistency regularization to assist with this optimization.

\subsection{Fidelity (Realism) Regularization}
\label{sec:r_feature}
We use the strong prior proposed in DeepInversion~\cite{yin2020dreaming} to guide the optimization towards natural images.
Specifically, we add $\mathcal{R_\text{fidelity}(\cdot)}$ to the loss function to steer $\mathbf{\hat x}$ away from unrealistic images with no discernible visual information: 
\begin{equation}
    \mathcal{R}_{\text{fidelity}}(\mathbf{\hat{x}}) = \alpha_{\text{tv}} \mathcal{R}_{\text{TV}}(\mathbf{\hat{x}}) + \alpha_{\ell_{2}} \mathcal{R}_{\ell_2}(\mathbf{\hat{x}}) + \alpha_\text{BN} \mathcal{R}_{\text{BN}}(\mathbf{\hat{x}}),
\label{eqn:r_prior}
\end{equation}
\noindent
where $\mathcal{R}_{\text{TV}}$ and $\mathcal{R}_{\ell_2}$ denote standard image priors~\cite{mahendran2015understanding, mordvintsev2015deepdream, nguyen2015deep} that penalize the total variance and $\ell_2$ norm of $\mathbf{\hat{x}}$, resp., with scaling factors $\alpha_{\text{tv}}$, $\alpha_{\ell_{2}}$. The key insight of DeepInversion resides in exploiting a strong prior in BN statistics:
\begin{eqnarray}
  \begin{aligned}
    \mathcal{R}_{\text{BN}}(\mathbf{\hat{x}}) 
    = & \sum_{l}|| \ \mu_{l}(\mathbf{\hat{x}}) - \text{BN}_{l}( \text{mean})||_2 + \\ 
    &\sum_{l}|| \ {\sigma^2_l}(\mathbf{\hat{x}}) - \text{BN}_{l}(\text{variance}) ||_2,
\end{aligned}
\label{eqn:fmap}
\vspace{-1mm}
\end{eqnarray}
\noindent
where $\mu_{l}(\mathbf{\hat{x}})$ and $\sigma_l^2(\mathbf{\hat{x}})$ are the batch-wise mean and variance estimates of feature maps corresponding to the $l^{\text{th}}$ convolutional layer. By enforcing valid intermediate distributions at all levels,  $\mathcal{R_\text{fidelity}(\cdot)}$ yields 
convergence towards realistic-looking solutions. 

\begin{figure}[t]
\centering
\resizebox{0.99\linewidth}{!}{
\begingroup
\renewcommand*{\arraystretch}{0.3}
\begin{tabular}{c|cccc}
\includegraphics[width=0.2\linewidth]{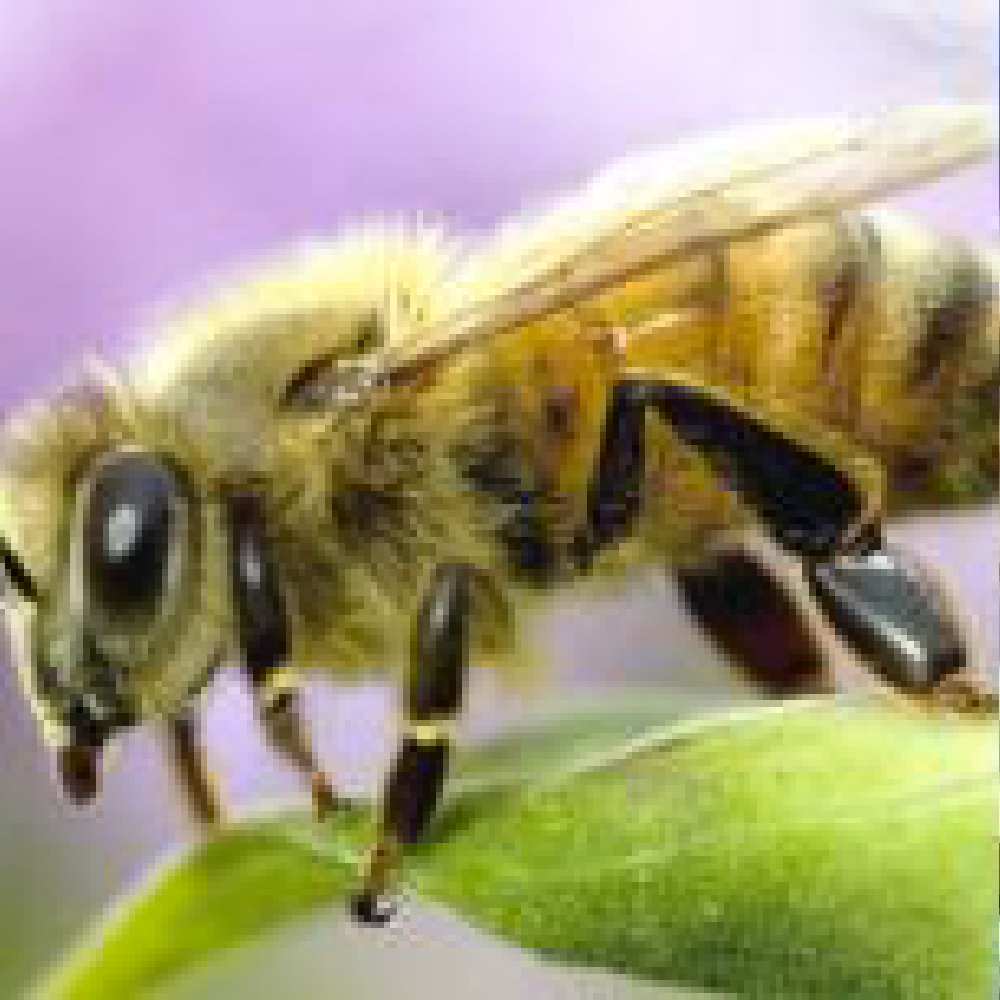}& \multicolumn{4}{c}{
\includegraphics[width=0.8\linewidth]{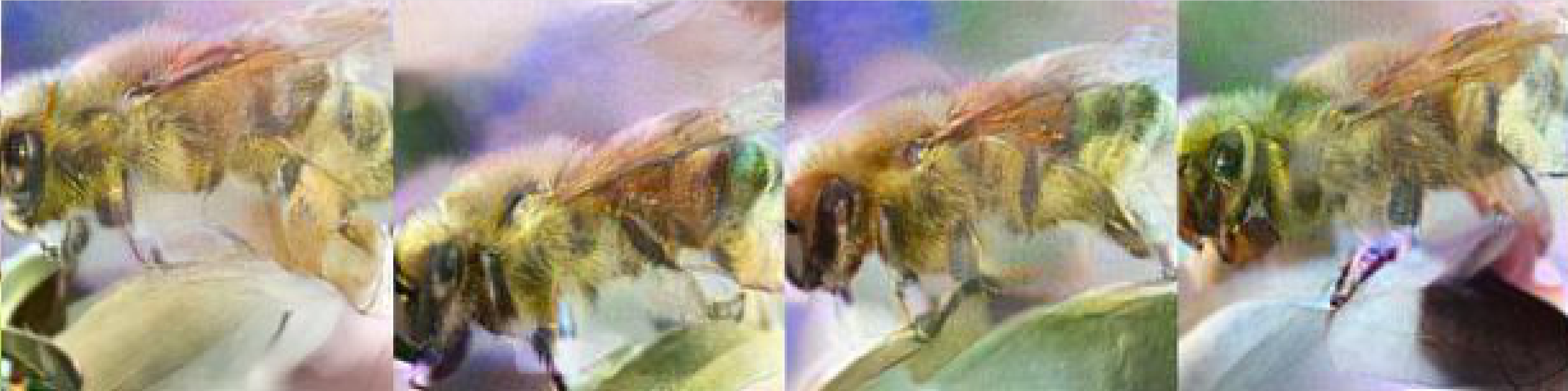}}\\
 \ \ \small{original} & \multicolumn{4}{c}{\small{results of independent optimization processes}}
\end{tabular}
\vspace{0.6mm}
\endgroup
}
\caption{Reconstruction variation in single-path optimization, focusing on  one target from a batch of size $8$. Optimizations follow the exact same loss hyperparameters, given only varying random seeds for pixel-wise initialization of $\mathbf{\hat{x}}$.} 
\label{fig:variation}
\vspace{-1mm}
\end{figure}

\subsection{Group Consistency Regularization}
\label{sec:r_group}
An additional challenge of gradient-based inversion lies in the exact localization of the target object, due to translational invariance of CNNs. Unlike an ideal scenario where optimization converges to one ground truth, we observe that when repeating the optimization with different seeds, \eg, as in Fig.~\ref{fig:variation}, 
each optimization process unveils a local minimum that allocates semantically correct image features at all levels, but differs from others -- images shift around the ground truth, focusing on slightly different details. During the forward pass, the existence of pooling layers, strided convolutions, and zero-padding, jointly causes spatial equivariance among the restored images, as also observed by Geiping \etal~\cite{geiping2020inverting}. A combination of the restored images from varying seeds, however, hints at the potential for a better restoration of the final image closer to the ground truth.

\begin{figure}[t]
\includegraphics[width=\linewidth]{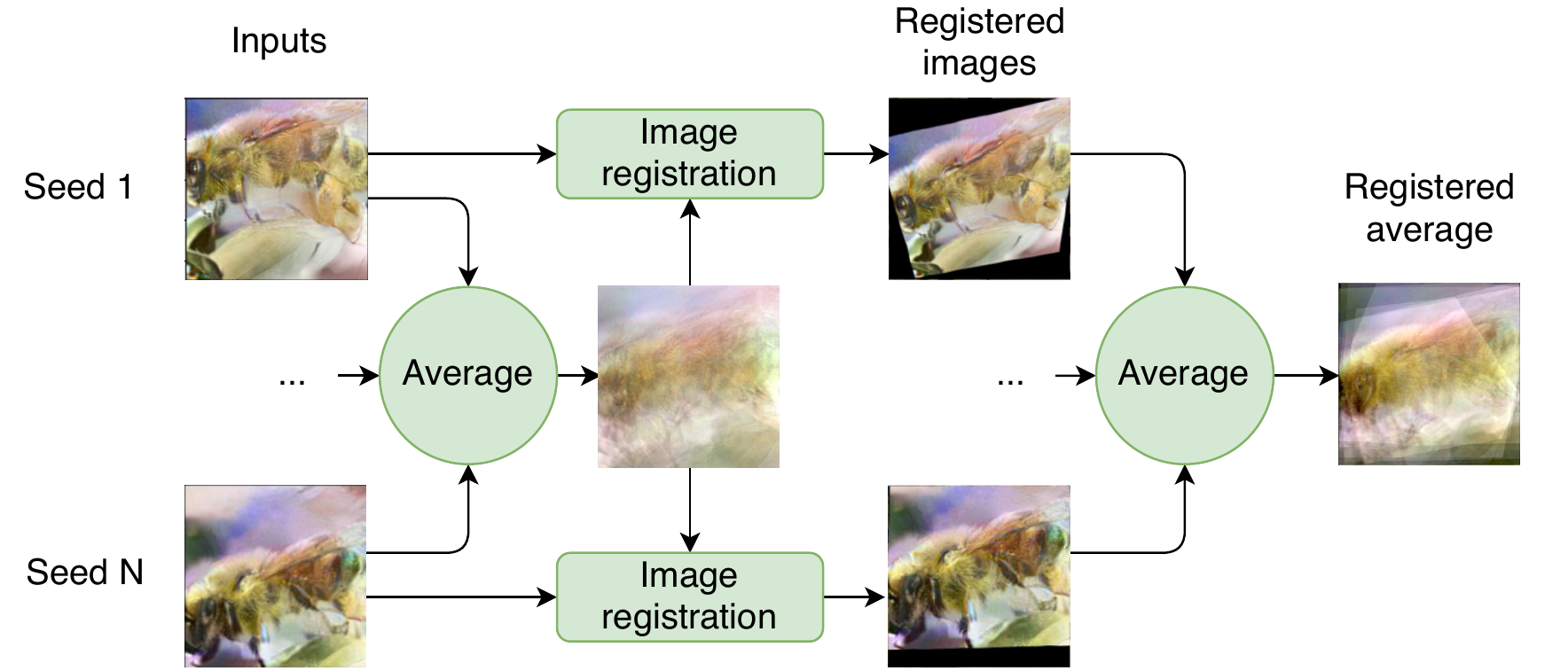}
\caption{Overview of group consistency regularization.}
\vspace{-2mm}
\label{fig:registration}
\end{figure}

We introduce a group consistency regularization term that exploits multiple seeds simultaneously in a joint optimization manner, as shown in Fig~\ref{fig:registration}. Intuitively, a joint exploration with multiple paths can expand and enlarge the search space during gradient descent.
However,  we have to regularize them to prevent too much divergence, at least in the final stages, given the search for a single target.
We optimize each input using the target Eqn.~\ref{eqn:main_error}. 
To facilitate information exchange, we regularize all the inputs
simultaneously
with a new group consistency regularization term:
\begin{equation}
\mathcal{R_\text{group}}(\mathbf{\hat{x}},  \mathbf{\hat{x}}_{g \in G}) = \alpha_{\text{group}} ||\mathbf{\hat{x}} -  \mathbb{E}(\mathbf{\hat{x}}_{g \in G})||_2,  
\label{eqn:r_group}
\end{equation}
\noindent
that jointly considers all the image candidates across all the seeds, and penalizes any candidate $\mathbf{\hat{x}}_g$ once it deviates away from the ``consensus'' image $\mathbb{E}(\mathbf{\hat{x}}_{g \in G})$ of the group.

One quick and intuitive option for $\mathbb{E}(\mathbf{\hat{x}}_{g \in G})$ is pixel-wise averaging. Though ``lazy'' as it seems, pixel-wise mean already leads to visual
improvements by mixing the information and feedback from all seeds in the group, as we will show later. To further explore the underlying transformations across seeds and create better consensus, we add in image registration to improve $\mathbb{E}(\mathbf{\hat{x}}_{g \in G})$: 
\begin{equation}
\mathbb{E}(\mathbf{\hat{x}}_{g \in G}) = \frac{1}{|G|} \sum_{g} \mathbf{F}_{\mathbf{\hat{x}}_{g} \rightarrow \frac{1}{|G|}\sum\limits_{g}{\mathbf{\hat{x}}_g}}(\mathbf{\hat{x}}_{g}).
\label{eqn:e_group}
\end{equation}
\noindent
This leads to our final group consistency regularization shown in Fig.~\ref{fig:registration}. We (i) first compute the pixel-wise mean within the candidate set of size $|G|$ as a coarse registration target, (ii) register each individual image towards the target via $\mathbf{F}(\cdot)$, 
and (iii) obtain the post-registered mean as the target for regularization. We use RANSAC-flow~\cite{shen2020ransac} for $\mathbf{F}(\cdot)$. 
As we will show later, group consistency regularization enables consistent improvements in recovery across various evaluation metrics, further closing the gap between reconstructed and original batches. 

\subsection{The Final Update}
Using all the above losses, we update the input in an iterative manner.
To further encourage exploration and diversity, we add pixel-wise random Gaussian noise in each update, inspired by the 
Langevin updates in energy-based models~\cite{du2019EBM, gao2018learning, grathwohl2019your}. Our final optimization steps are:
\begin{eqnarray}
\label{eq:final_update}
\Delta_{\mathbf{\hat{x}}^{(t)}} & \leftarrow & \nabla_\mathbf{\hat{x}} \big(\mathcal{L_\text{grad}}(\mathbf{\hat{x}}^{{(t-1)}}, \nabla \mathbf{W}) +  \mathcal{R_\text{aux}}(\mathbf{\hat{x}}^{(t-1)}) \big) \nonumber \\
\eta & \leftarrow & \mathcal{N}(0,\mathcal{I}) \nonumber \\
\mathbf{\hat{x}}^{(t)} & \leftarrow & \mathbf{\hat{x}}^{(t-1)} + \lambda(t) \Delta_{\mathbf{\hat{x}}^{(t)}}  + \lambda(t) \alpha_n \eta \nonumber
\end{eqnarray}
where $\Delta_{\mathbf{\hat{x}}^{(t)}}$ corresponds to an optimizer update, $\eta$ denotes randomly sampled noise to encourage exploration, $\lambda(t)$ is the learning rate, and $\alpha_n$ re-scales the finally added noise.

\section{Experiments}
\label{sec:exp}
We evaluate our method for the classification task on the large-scale 1000-class ImageNet ILSVRC 2012 dataset~\cite{deng2009imagenet} at $224\times 224$ pixels. We first perform a number of ablations to evaluate the contribution of each component of our method. Then, we show the success of GradInversion and compare with prior art. 
Finally, we increase the batch size to explore the limits of gradient inversion.

\noindent
\textbf{Implementation details.} In all cases, image pixels are initialized $i.i.d.$ from Gaussian noise of $\mu=0$ and $\sigma=1$. We primarily focus on the ResNet-50 architecture for the classification task, pre-trained with MOCO V2 and fine-tuned only the classification layer, achieving $71.0\%$ top-1 accuracy on ImageNet\footnote{Based on \url{https://github.com/facebookresearch/moco}. MOCO V2 (Chen \etal~\cite{chen2020improved}) enhances MOCO (He \etal~\cite{he2020momentum}) with SimCLR (Chen \etal~\cite{chen2020simple}) and reports ImageNet top-1 accuracy at $71.1\%$~\cite{chen2020improved}.}. We observe that stronger feature extraction leads to better restoration 
as compared to the default pre-trained PyTorch model, and shallower network architectures (ResNet-18). 
We use Adam for optimization with a $0.1$ learning rate with cosine learning rate decay, and $50$ iterations as warm up. We use $\alpha_{\text{tv}}=1\cdot10^{-4}, \alpha_{\ell_2}=1\cdot10^{-6}, \alpha_{\text{BN}}=0.1,  \alpha_{\text{G}}=0.001, \alpha_{\text{group}}=0.01, \alpha_n=0.2$ as loss scaling constants. 
For feature distribution regularization, we primarily focus on the case when BN statistics of the target batch is jointly provided with the gradients as commonly required in distributed learning for global BN updates~\cite{liu2018path,zhang2018context,zhao2017pyramid}. We also analyze regularization towards network BN means and variances - averaged over dataset, they offer proxy for single batch statistics. We synthesize image batches of resolution $224\times224$ using NVIDIA V100 GPUs and automatic-mixed precision (AMP)~\cite{micikevicius2017mixed} acceleration. Optimization of each batch consumes $20$K optimization iterations.

\noindent
\textbf{Evaluation metrics.} We present visual comparisons of images obtained under different settings and evaluate three quantitative metrics for image similarity. To account for pixel-wise mismatch, we compute: (i) the cosine similarity in FFT$_{\text{2D}}$ frequency response, (ii) post-registration PSNR, and (iii) LPIPS perceptual similarity score~\cite{zhang2018perceptual} between reconstruction and original images.
\vspace{-0.5mm}

\subsection{Ablation Studies}
\vspace{-1mm}
\subsubsection{Label restoration}
\vspace{-1mm}
We first restore labels from the gradients of the fully connected layer. Table~\ref{tab:label_restore} summarizes the averaged label restoration accuracy on ImageNet training and validation sets, given $10$K randomly drawn samples divided into varying batch sizes. In a zero-shot method, GradInversion restores original labels accurately, improving upon prior art~\cite{zhao2020idlg}. 


\begin{table}[!t]
\centering
\resizebox{.6\linewidth}{!}{
\begin{tabular}{ccccc}
\toprule
 \multirow{2}{*}{\textbf{Batch}} & \multicolumn{4}{c}{\textbf{Label Restoration Accuracy ($\%$)}}\\
 \cmidrule{2-5}
 & \multicolumn{2}{c}{Training Set} & \multicolumn{2}{c}{Validation Set} \\ 
\textbf{size} & \cite{zhao2020idlg}$^\dagger$ & \textbf{ours} & \cite{zhao2020idlg}$^\dagger$ & \textbf{ours} \\
\midrule
$1$     & $100.0$ & $100.0$ & $100.0$ & $100.0$ \\
$8$     & $95.89$ & $99.56$ & $96.08$ & $99.47$ \\
$32$    & $89.88$ & $99.29$ & $90.32$ & $99.19$ \\
$64$    & $84.51$ & $98.79$ & $82.27$ & $98.21$ \\
$96$    & $80.53$ & $97.88$ & $82.13$ & $98.11$ \\
\bottomrule
\end{tabular}}
\caption{Average restoration accuracy over $10$K random samples of different batch size from the ImageNet training/validation sets without label repeats. $^\dagger$: the original method~\cite{zhao2020idlg} only works for single image - we extend it by adopting its sum rule for Eqn.~\ref{eqn:sbatch} and then show improvements.}
\label{tab:label_restore}
\end{table}


\begin{table}[!t]
\centering
\resizebox{.98\linewidth}{!}{
\begin{tabular}{lcccc}
\toprule
\multirow{2}{*}{\textbf{Obj. Function}}  & \multirow{2}{*}{$\mathbf{\mathcal{L}_\text{grad}(\mathbf{\hat{x}^*}; \nabla \mathbf{W})}$} & \multicolumn{3}{c}{\textbf{Distance to Original Images}}\\
\cmidrule{3-5}
& & FFT$_\text{2D}$ $\downarrow$ & PSNR $\uparrow$ & LPIPS $\downarrow$  \\
\midrule
 \ \ \ \ \ $\mathcal{N}(0, \mathcal{I})$ & $8.625$ & $0.706$ & $9.964$  & $1.351$  \\
\midrule
 \ \ \ \ \ $\mathcal{L_\text{grad}}$ & $4.190$ & $0.404$ & $10.753$  & $0.919$ \\
 \ \ + $\mathcal{R_\text{fidelity}}$ & $3.206$ & $0.279$ & $12.058$  & $0.655$ \\
 \ \ + $\mathcal{R_\text{group.lazy}}$ & $2.918$    & $0.233$ & $12.261$  & $0.578$  \\
 \ \ + $\mathcal{R_\text{group.reg}}$ & $\mathbf{2.685}$ & $\mathbf{0.175}$ & $\mathbf{12.929}$  & $\mathbf{0.484}$ \\
\bottomrule
\end{tabular}}
\vspace{1mm}

\resizebox{0.99\linewidth}{!}{
\begingroup
\begin{tabular}{c|cccc}
\centering
\includegraphics[width=0.2\linewidth]{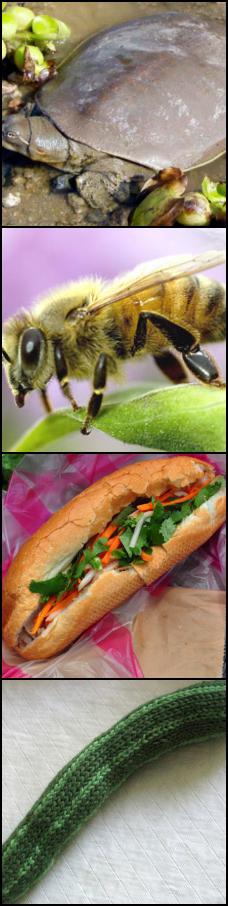} &
\includegraphics[width=0.2\linewidth]{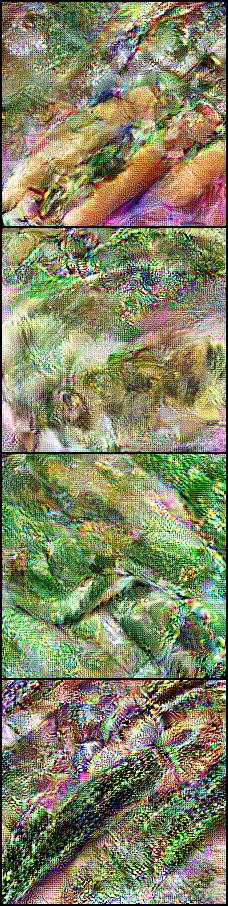} &
\includegraphics[width=0.2\linewidth]{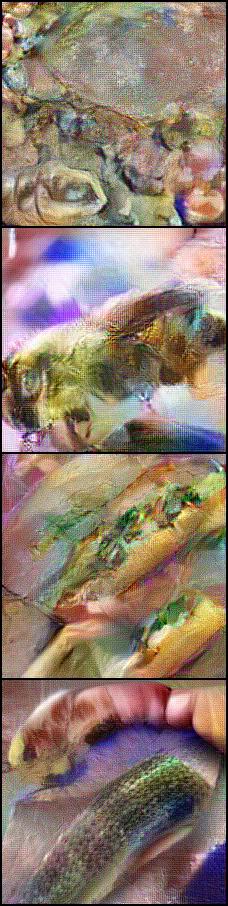} &
\includegraphics[width=0.2\linewidth]{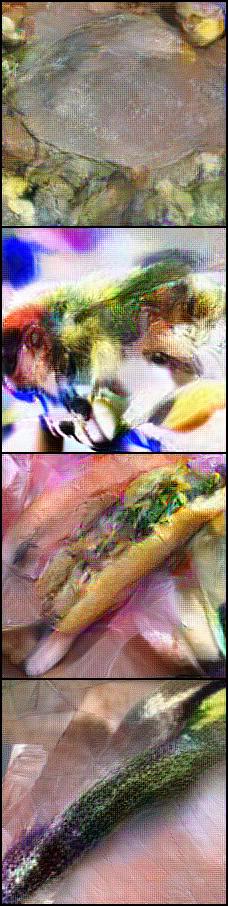} &
\includegraphics[width=0.2\linewidth]{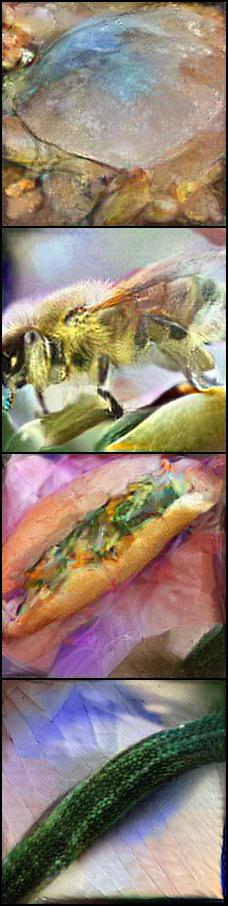} \\
$\mathbf{x^*}$ & $\mathcal{L_\text{grad}}$ & $+\mathcal{R_\text{fidelity}}$ & $+\mathcal{R_\text{group.lazy}}$ & $+\mathcal{R_\text{group.reg}}$
\end{tabular}
\endgroup
}
\caption{Ablation study when each proposed loss to optimization objective function - quantitative (up) and qualitative (bottom) comparison. Original batch contains $8$ samples - we show $4$ samples visually here amid space limit, see Appendix for entire batch. 
} 
\vspace{0.2mm}
\label{tb:loss_ablation_table}
\end{table}

\vspace{-1mm}

\subsubsection{Batch reconstruction}
We next gradually add each proposed loss to the optimization process. Here we focus on a batch of $8$ images for algorithm ablations before expanding towards a larger batch size. We summarize results in Table~\ref{tb:loss_ablation_table} and discuss insights next:


\noindent
\textbf{Adding $\mathcal{L}_{grad}$.} 
We find $\ell_2$ loss outperforms cosine similarity~\cite{geiping2020inverting} for gradient matching - see Appendix for details. 
Reconstructed images remain noisy. Partial original features emerge, but leak among images within the batch.

\noindent
\textbf{Adding $\mathcal{R}_{\text{fidelity}}$.} Adding fidelity regularization immediately improves image quality. 
Conditioned on image prior, gradient inversion starts to allocate visual details towards individual images, enabling both visual and quantitative improvements in Table~\ref{tb:loss_ablation_table}. 

\begin{table}[!t]
\centering
\resizebox{.78\linewidth}{!}{
\begin{tabular}{lcccc}
\toprule
\multirow{2}{*}{\textbf{Model}}  & \multicolumn{4}{c}{\textbf{Distance to Original Images}}\\
\cmidrule{2-5}
& FFT$_\text{2D}$ $\downarrow$ & PSNR $\uparrow$  & LPIPS $\downarrow$  \\
\midrule
ResNet-50 (MOCO V2)   & $0.175$ & $12.929$  & $0.484$ \\
ResNet-50 (standard)  & $0.204$ & $11.771$  & $0.584$ \\
ResNet-18 (standard)  & $0.218$ & $10.729$  & $0.693$ \\
\bottomrule
\end{tabular}}
\caption{Reconstruction under varying feature extraction strength.}
\label{tab:nomoco}
\vspace{-1mm}
\end{table}

\begin{figure*}[t]
\centering

\resizebox{1.\linewidth}{!}{
\begingroup
\renewcommand*{\arraystretch}{0.3}
\begin{tabular}{c}
\includegraphics[width=1.1\linewidth]{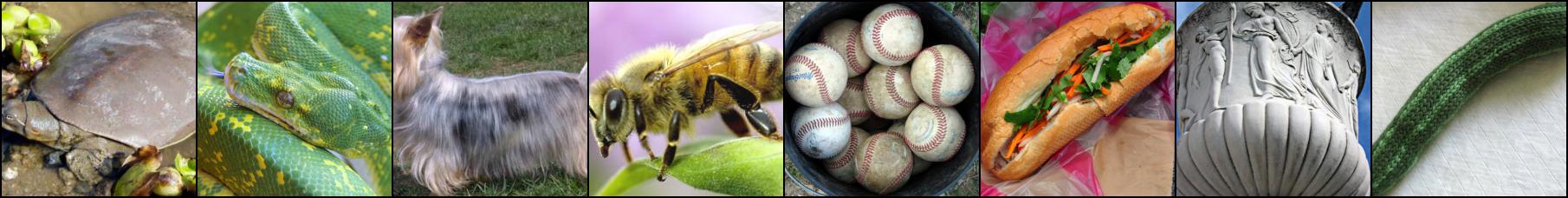} \\
 \textbf{\small{Original batch - ground truth}}\\
 \midrule


 \includegraphics[width=1.1\linewidth]{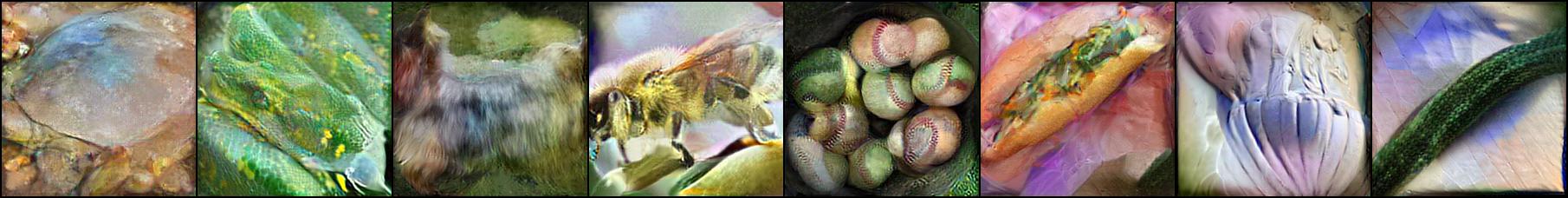} \\
 \small{GradInversion \textbf{(Ours)} - LPIPS $\downarrow$: \textbf{0.484}}
 \vspace{1mm}\\ 
 
 \includegraphics[width=1.1\linewidth]{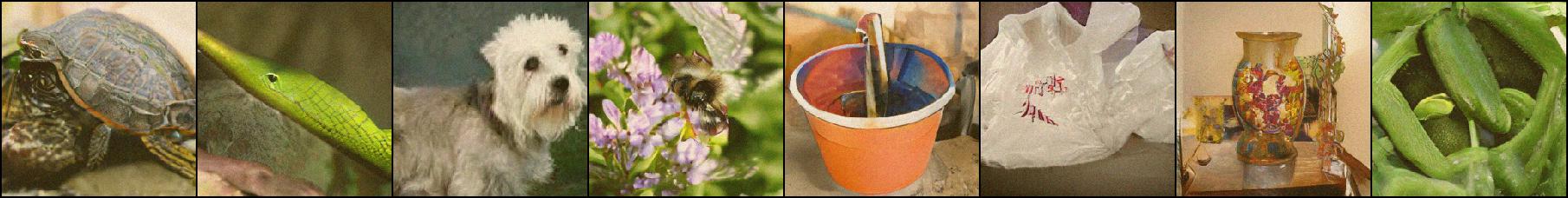} \\
  \small{Latent Projection (Karras \etal CVPR'20~\cite{karras2020analyzing}) of BigGAN-deep (Brock \etal ICLR'19~\cite{brock2018biggan}) for Gradient Matching - LPIPS $\downarrow$: 0.732}
\vspace{1mm}\\
  
  \includegraphics[width=1.1\linewidth]{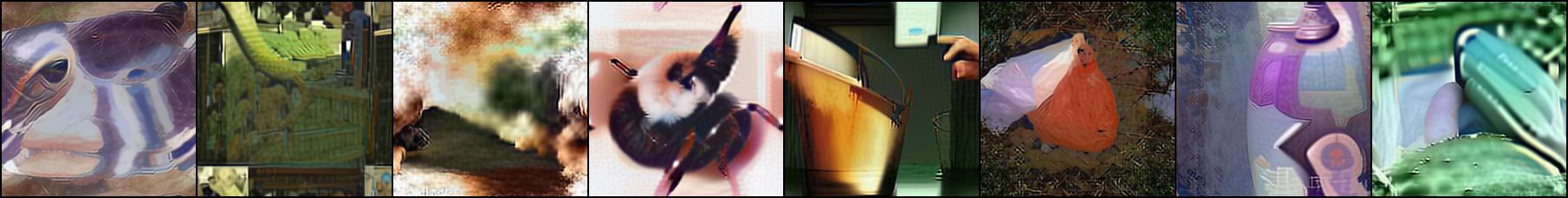} \\
  \small{DeepInversion (Yin \etal CVPR'20~\cite{yin2020dreaming}) - LPIPS $\downarrow$: 0.728}
    \vspace{1mm}\\
  
  \includegraphics[width=1.1\linewidth]{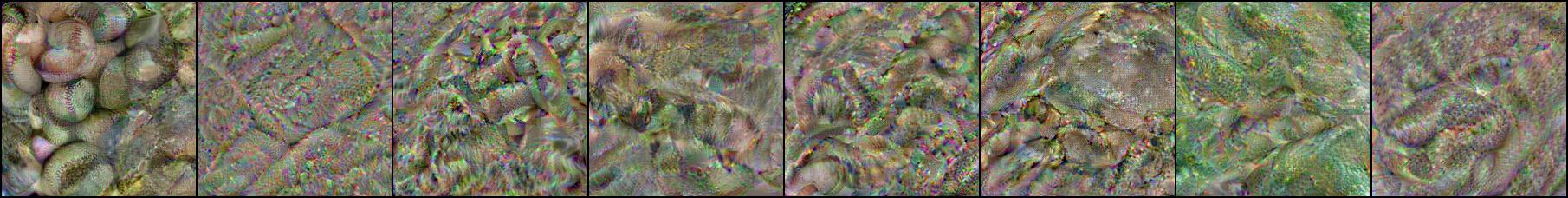} \\ 
  \small{Inverting Gradients (Geiping \etal NeurIPS'20~\cite{geiping2020inverting}) - LPIPS $\downarrow$: 0.749} 
    \vspace{1mm} \\
  
 \includegraphics[width=1.1\linewidth]{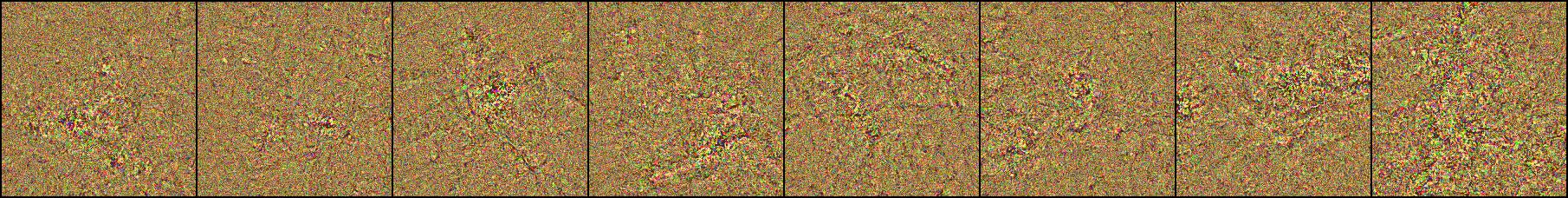} \\
  \small{Deep Gradient Leakage (Zhu \etal NeurIPS'19~\cite{zhu2019deep}) - LPIPS $\downarrow$: 1.319}  
  \vspace{1mm}\\
\end{tabular}
\endgroup
}
\caption{ImageNet batch gradient inversion for ResNet-50 visual comparison with state-of-the-art methods. GradInversion labels rearranged in ascending order to match ground truth after label restoration at $100\%$ accuracy. Best viewed in color.}
\vspace{-1.1mm}
\label{fig:against_sota}
\end{figure*}

\noindent
\textbf{Adding $\mathcal{R}_{\text{group}}$.} 
Group consistency regularization further improves reconstruction. For this analysis, we use $8$ random seeds, each determining a Gaussian initialization of inputs and the associated pixel-wise perturbations. All seeds are jointly optimized, compatible with standard multi-node training pipeline that supports synchronization only for $\mathbb{E}(\mathbf{\hat{x}}_{g \in G})$ computation.
For better insights, we next compare the choice of (i) ``lazy'' pixel-wise mean, and (ii) registration-augmented mean as the regularization target.

\noindent
\textbf{a) Lazy regularization.} 
We observe ``lazy'' pixel-wise mean as regularization target already brings in performance improvements. Though not yet accommodating for inter-seed variation, pixel-wise mean hints on correct ``perceived'' positions of the target objects. Objects start to emerge at correct positions with improved orientations. 

\noindent
\textbf{b) Registration enhancement.} We then add in registration to exploit consensus among candidates. 
We start registration after $5$K initial optimization iterations to allow for sufficient feature emergence, then iterate every $100$ iterations. Ideally, each candidate shall be registered to its original image for the best spatial adjustment. While given no such access, registration to pixel-wise mean turns out to be effective. The final registration-based regularization helps close the remaining gap - it improves all evaluation metrics in Table~\ref{tb:loss_ablation_table}. At this stage, GradInversion accurately allocates detailed original contents to \textit{individual} images, from  \textit{averaged} gradients. 

\noindent
\textbf{Inverting different networks.}
We observe that gradients from a 
stronger feature extractor 
leak more information - see a quick comparison in Table~\ref{tab:nomoco}. Self-supervised pretraining of ResNet-50 leads to the best image reconstruction, when compared to a standard training recipe of the same ResNet-50 architecture, and a weaker ResNet-18. We continue our analysis with ResNet-50 MOCO V2 to study the limits of batch reconstruction under gradient inversion.
\vspace{-1mm}

\subsection{Comparison with the state-of-the-art} 
\begin{figure}[t]
\centering
\resizebox{0.75\linewidth}{!}{
\begingroup
\renewcommand*{\arraystretch}{0.3}
\begin{tabular}{ccc}
\includegraphics[width=0.33\linewidth,clip,trim=5px 0 0 4px]{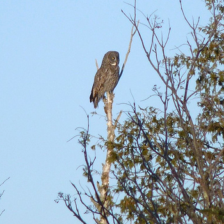} &
\includegraphics[width=0.33\linewidth,clip,trim=5px 0 0 4px]{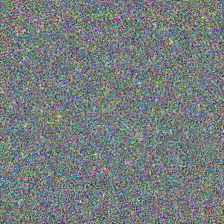} &
\includegraphics[width=0.33\linewidth,clip,trim=5px 0 0 4px]{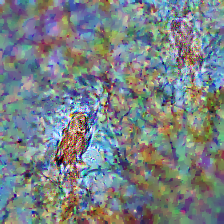}
\\
Original & Zhu \etal~\cite{zhu2019deep} & Geiping \etal~\cite{geiping2020inverting}
\\
\end{tabular}
\endgroup
}

\resizebox{0.5\linewidth}{!}{
\begingroup
\renewcommand*{\arraystretch}{0.3}
\begin{tabular}{cc}
\includegraphics[width=0.33\linewidth,clip,trim=4px 0 0 4px]{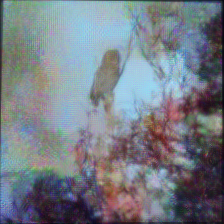} &
\includegraphics[width=0.33\linewidth,clip,trim=5px 0 0 4px]{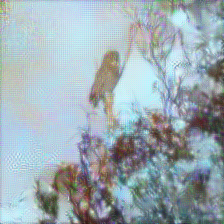}
\\
(d) \textbf{Ours} - BN$_{\text{approx.}}$ & (e) \textbf{Ours} - BN$_{\text{exact}}$\\
\end{tabular}
\endgroup
}
\caption{Comparison with prior art on ResNet-50 (ImageNet) gradient inversion at batch size 1 for a ``challenging'' sample from~\cite{geiping2020inverting}. 
}
\label{fig:bs1_comparison}
\end{figure}

We next compare with prior art on the batch size of 8 images with 224x224px. We summarize both qualitative (Fig.~\ref{fig:against_sota}) and quantitative results (Table~\ref{tab:against_sota}). We compare with three viable methods for image synthesis: 

\noindent\textbf{(i) Gradient inversion~\cite{geiping2020inverting, zhu2019deep}:} We first compare with prior model inversion methods for gradient matching: (i) deep gradient leakage method by Zhu \etal~\cite{zhu2019deep} 
and (ii) federated gradient inversion by Geiping \etal~\cite{geiping2020inverting}. 
We first extend both techniques towards ImageNet batch restoration following the authors' public open-sourced repository~\cite{zgeiping_github,zzhu_github}. 
For an additional fair comparison, we also compare with both methods at batch size one in Fig.~\ref{fig:bs1_comparison}, and show notable fidelity and localization improvements. 



\noindent\textbf{(ii) DeepInversion~\cite{yin2020dreaming}:} We also analyze performance improvements over the baseline DeepInversion method that synthesizes images conditioned on ground-truth labels. 

\noindent\textbf{(iii) GAN latent space projection~\cite{karras2020analyzing}:} We finally compare with the GAN-based latent code optimization method. We applied latent code projection as in StyleGAN$2$ ~\cite{karras2020analyzing} for BigGAN-deep generator~\cite{brock2018biggan} at resolution $256\times256$. Given no access to original images for projection loss~\cite{karras2020analyzing}, we base the target loss on $\ell_2$ distances between synthesized and ground truth gradients.


GradInversion outperforms prior art both visually (Fig.~\ref{fig:against_sota}) and numerically (Table~\ref{tab:against_sota}). Without label restoration, a joint optimizing to seek for image-label pairs~\cite{zhu2019deep} struggles to converge on ImageNet, as also observed by~\cite{geiping2020inverting} even at batch size one. Total variation prior and magnitude-invariant loss as in~\cite{geiping2020inverting} help improve reconstruction, but remain too weak to guide optimization towards ground truth. The DeepInversion~\cite{yin2020dreaming} baseline improves image fidelity as expected, but inverts images with little observable links to the original batch. Projection onto BigGAN's latent space offers a balance between image fidelity and restored details, but falls short under a weaker guidance from original gradients rather than original images, as projection to latent space is NP-hard~\cite{lei2019inverting} and misses visual details~\cite{karras2020analyzing}.

\begin{table*}[!t]
\centering
\resizebox{\linewidth}{!}{
\begin{tabular}{lccccc}
\toprule
\multirow{2}{*}{\textbf{Method}} & \multicolumn{2}{c}{\textbf{Requirements}} & \multicolumn{3}{c}{\textbf{Distance to Original Images}}\\
\cmidrule{2-3}\cmidrule{4-6}
& $\mathbf{y}^*$ & GAN & FFT$_\text{2D}$ $\downarrow$ & PSNR $\uparrow$ & LPIPS $\downarrow$  \\
\midrule
Noise $\mathcal{N}(0, I)$ (starting point) & - & - & $0.706$ & $9.964$ & $1.351$ \\
\midrule

Latent projection (Karras \etal CVPR'20~\cite{karras2020analyzing}) of BigGAN-deep (Brock \etal ICLR'19~\cite{brock2018biggan}) & \checkmark & \checkmark & $0.275$ &  $10.149$ & $0.722$ \\

DeepInversion (Yin \etal CVPR'20~\cite{yin2020dreaming})   & \checkmark & - & $0.238$ & $10.131$ & $0.728$ \\
Inverting Gradients (Geiping \etal NeurIPS'20~\cite{geiping2020inverting})          & \checkmark & - & $0.355$ & $11.703$ & $0.749$ \\
Deep Gradient Leakage (Zhu \etal NeurIPS'19~\cite{zhu2019deep})                  & - & -  & $0.602$ & $10.252$ & $1.319$ \\
\midrule
GradInversion - BN$_\text{approx.}$ \textbf{(ours)} & - & - & $0.232$ & $11.235$ & $0.633$  \\
GradInversion - BN$_\text{exact}$ \ \ \ \textbf{(ours)} & - & - & $\mathbf{0.175}$ & $\mathbf{12.929}$ & $\mathbf{0.484}$ \\
\bottomrule
\end{tabular}}
\caption{Comparison of GradInversion with state-of-the-art methods for ResNet-50 gradient inversion on ImageNet1K. BN$_\text{approx.}$ denotes regularizing towards BN statistics in the network learnt from the original dataset; BN$_\text{exact}$ denotes the BN statistics of target batch shared (or leaked) in distributed setup for global BN updates, 
\eg, Synchronized Batch Normalization~\cite{zhang2018context}.}
\vspace{-2mm}
\label{tab:against_sota}
\end{table*}

\subsection{Effect of scaling up the batch size}
\begin{figure}[t]
\centering
\resizebox{0.97\linewidth}{!}{
\begingroup
\renewcommand*{\arraystretch}{0.3}
\begin{tabular}{c|ccc}
\includegraphics[width=0.3\linewidth,clip,trim=5px 0 0 4px]{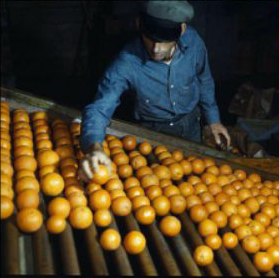} &
\includegraphics[width=0.3\linewidth,clip,trim=5px 0 0 4px]{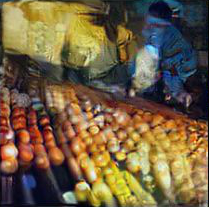} &
\includegraphics[width=0.3\linewidth,clip,trim=5px 0 0 4px]{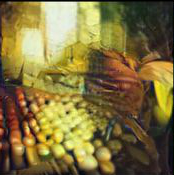} &
\includegraphics[width=0.3\linewidth,clip,trim=5px 0 0 4px]{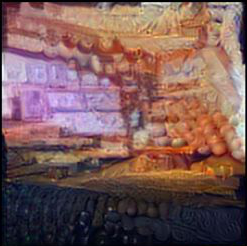}
\\
\multirow{2}{*}{\textbf{Original}} & batch size $4$ & batch size $16$ & batch size $48$ 
\vspace{0.5mm} \\
 & \multicolumn{3}{c}{\textbf{Restored}} \\
\end{tabular}
\endgroup
}
\caption{Reduced amount of restored original visual features as batch size increases.} 
\vspace{-3mm}
\label{fig:vanishing_feature}
\end{figure}

\begin{figure}[t]
\centering

\resizebox{\linewidth}{!}{
\begingroup
\renewcommand*{\arraystretch}{0.3}
\begin{tabular}{ccc}
\includegraphics[width=0.45\linewidth,clip,trim=5px 0 0 4px]{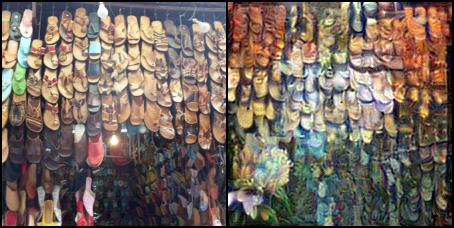} &
\includegraphics[width=0.45\linewidth,clip,trim=5px 0 0 4px]{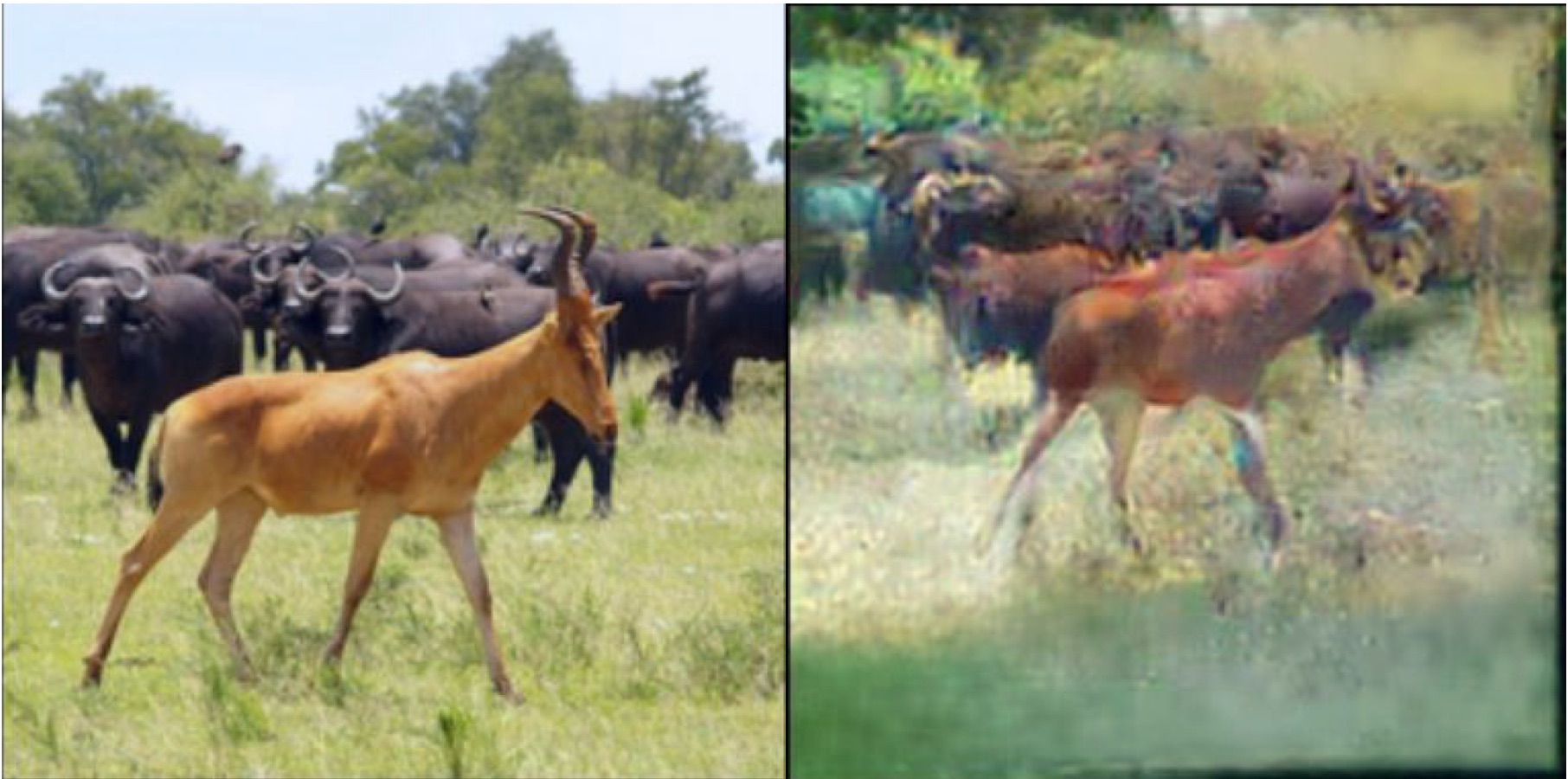} &
\includegraphics[width=0.45\linewidth,clip,trim=5px 0 0 4px]{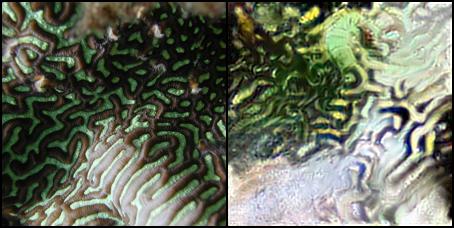}
\\
\multicolumn{3}{c}{(a) details restored}
\end{tabular}
\endgroup
}

\resizebox{\linewidth}{!}{
\begingroup
\renewcommand*{\arraystretch}{0.3}
\begin{tabular}{ccc}
\includegraphics[width=0.45\linewidth,clip,trim=5px 0 0 4px]{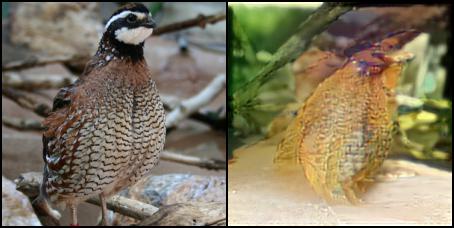} &
\includegraphics[width=0.45\linewidth,clip,trim=5px 0 0 4px]{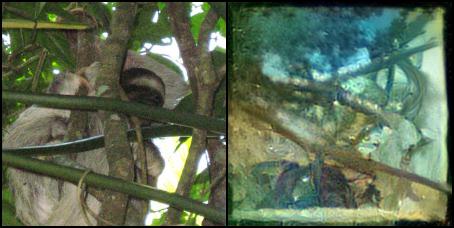} &
\includegraphics[width=0.45\linewidth,clip,trim=5px 0 0 4px]{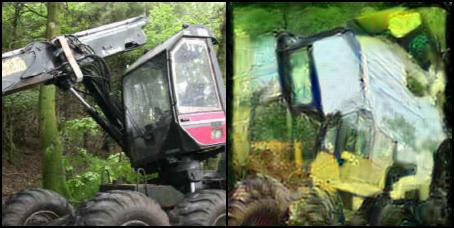} \\
\multicolumn{3}{c}{(b) semantics restored}
\end{tabular}
\endgroup
}

\resizebox{\linewidth}{!}{
\begingroup
\renewcommand*{\arraystretch}{0.3}
\begin{tabular}{ccc}
\includegraphics[width=0.45\linewidth,clip,trim=5px 0 0 4px]{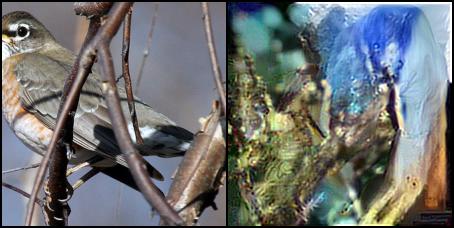} &
\includegraphics[width=0.45\linewidth,clip,trim=5px 0 0 4px]{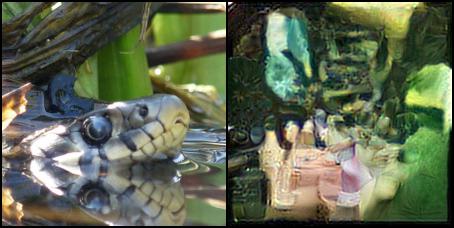} &
\includegraphics[width=0.45\linewidth,clip,trim=5px 0 0 4px]{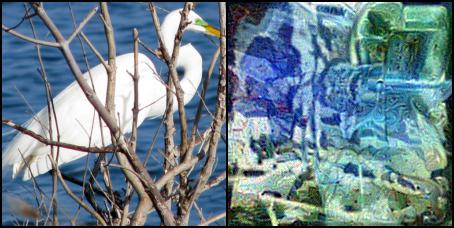} \\
\multicolumn{3}{c}{(c) no visual information}
\end{tabular}
\endgroup
}
\caption{Varying level of information leakage at batch size $48$ on ImageNet validation set. Each block containing a pair of (left) original sample and its (right) reconstruction by GradInversion. }
\vspace{-0.5mm}
\label{fig:bs48_imgs}
\end{figure}

We next increase the batch size. Our current analysis scales up to batch size $48$ using a $32$GB NVIDIA V100 GPU. As shown in Fig.~\ref{fig:vanishing_feature}, the amount of recoverable image content gradually decreases as batch size increases. As expected, more averaging of gradient information in a batch better protects privacy of an individual image. Surprisingly, GradInversion still unveils a decent amount of original visual information at batch size $48$, and sometimes a viable complete reconstruction, as shown in Fig.~\ref{fig:bs48_imgs}. 



\begin{figure}[!t]
\centering
\includegraphics[width=\columnwidth]{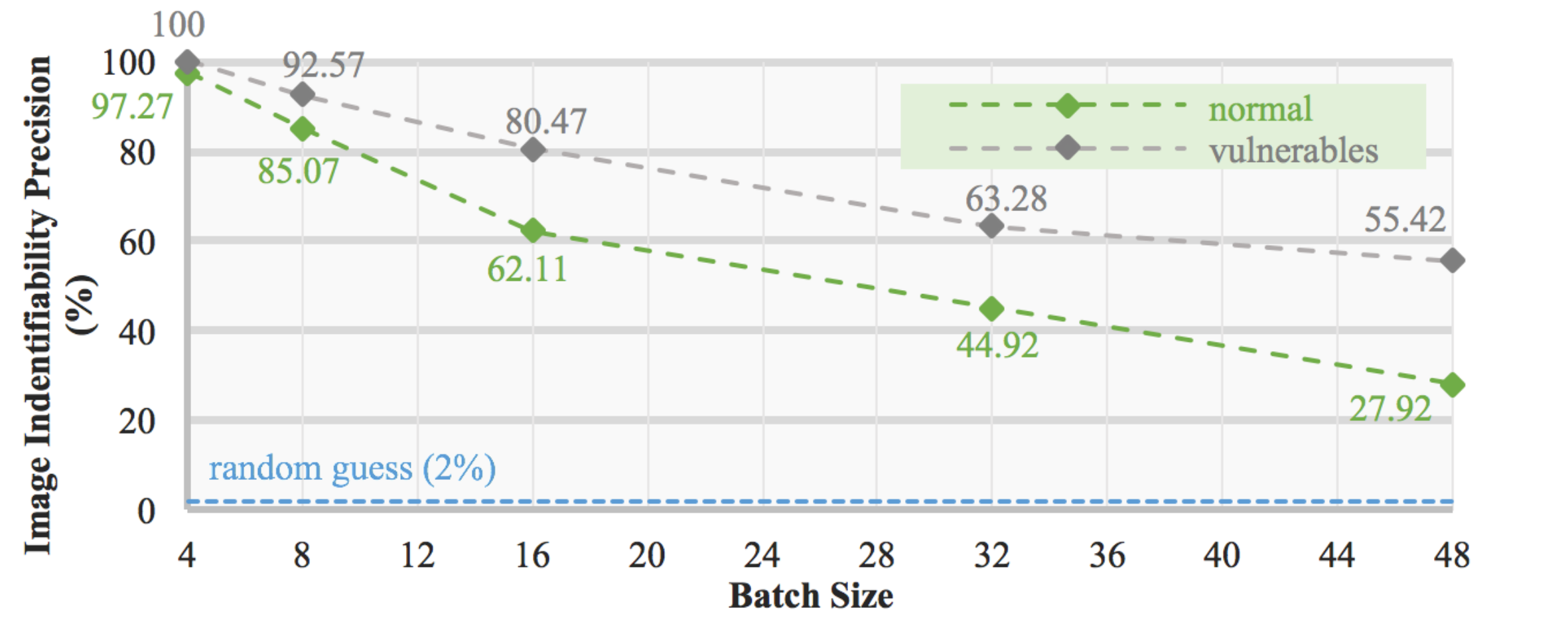}
\caption{The Image Indentifiability Precision (IIP)  curve of GradInversion on ImageNet validation set, as a function of increasing batch size. Each point averaged per $256$ randomly selected samples of varying batch sizes ($240$ samples for batch size 48). Nearest neighbors measured in avgpool feature space cosine similarity. 
}
\vspace{-0.8mm}
\label{fig:fetching_acc}
\end{figure}

\noindent
\textbf{Image Identifiablity Precision (IIP).} 
We formulate a new score that measures the amount of ``image-specific'' features revealed by gradient inversion. Intuitively, this measures how easy it is to identify a particular image, given only its reconstruction, among all its similar peers in the original dataset. Quantifiably, we calculate the fraction of exact matches between an original image and the nearest neighbor to its reconstruction. The resulting metric, referred to as \textit{Image Identifiability Precision (IIP)}, evaluates gradient inversion strength across varying batch sizes.
Fig.~\ref{fig:fetching_acc} plots the IIP curve for GradInversion. As expected, reconstruction efficacy gradually decreases as batch size increases, as also seen in 
Fig.~\ref{fig:vanishing_feature}. We make a surprising observation that many samples ($\sim 28\%$) can be correctly identified even after averaging gradient from 48 images.

\noindent
\textbf{The Vulnerable Population.} We empirically observe a positive correlation between reconstruction efficacy and gradient magnitude. Delving deeper into this observation, we identify a new set of images that are more ``vulnerable'' to leak information under GradInversion. To this end, we identify one image per ImageNet class category whose gradient $\ell_2$ norm is the largest within that class folder. When compared to random images in Fig.~\ref{fig:fetching_acc}, batches sampled from such ``vulnerable population'' increases the IIP by large margins, nearly doubled at batch size $48$. This advocates for careful attention to such vulnerable samples before gradient sharing.





\section*{Conclusions}
We introduced GradInversion to reconstruct individual images in a batch, given averaged gradients. 
We showed that the assumption of privacy when sharing gradients from deep networks on complex datasets even at large batch sizes, does not hold. 
This offers new insights into the development of privacy-preserving deep learning frameworks. 

It can also be fruitful to study the underlying mechanism of information transfer that enables original data recovery from gradients.
We hope that future work can study vulnerabilities of aggregration-based federated learning~\cite{bonawitz2017practical}, as well as further strengthen them to prevent inversion. 



{\small
\bibliographystyle{ieee}
\bibliography{bib}

\begin{thebibliography}{10}\itemsep=-1pt

\bibitem{bonawitz2019towards}
K.~Bonawitz, H.~Eichner, W.~Grieskamp, D.~Huba, A.~Ingerman, V.~Ivanov,
  C.~Kiddon, J.~Kone{\v{c}}n{\`y}, S.~Mazzocchi, H.~B. McMahan, et~al.
\newblock Towards federated learning at scale: {S}ystem design.
\newblock {\em SysML}, 2019.

\bibitem{bonawitz2017practical}
K.~Bonawitz, V.~Ivanov, B.~Kreuter, A.~Marcedone, H.~B. McMahan, S.~Patel,
  D.~Ramage, A.~Segal, and K.~Seth.
\newblock Practical secure aggregation for privacy-preserving machine learning.
\newblock In {\em CCS}, 2017.

\bibitem{brisimi2018federated}
T.~S. Brisimi, R.~Chen, T.~Mela, A.~Olshevsky, I.~C. Paschalidis, and W.~Shi.
\newblock Federated learning of predictive models from federated electronic
  health records.
\newblock {\em IJMI}.

\bibitem{brock2018biggan}
A.~Brock, J.~Donahue, and K.~Simonyan.
\newblock Large scale {GAN} training for high fidelity natural image synthesis.
\newblock In {\em ICLR}, 2019.

\bibitem{cai2020zeroq}
Y.~Cai, Z.~Yao, Z.~Dong, A.~Gholami, M.~W. Mahoney, and K.~Keutzer.
\newblock {Z}ero{Q}: {A} novel zero shot quantization framework.
\newblock In {\em CVPR}, 2020.

\bibitem{chen2019data}
H.~Chen, Y.~Wang, C.~Xu, Z.~Yang, C.~Liu, B.~Shi, C.~Xu, C.~Xu, and Q.~Tian.
\newblock Data-free learning of student networks.
\newblock In {\em ICCV}, 2019.

\bibitem{chen2020simple}
T.~Chen, S.~Kornblith, M.~Norouzi, and G.~Hinton.
\newblock A simple framework for contrastive learning of visual
  representations.
\newblock {\em arXiv preprint arXiv:2002.05709}, 2020.

\bibitem{chen2020improved}
X.~Chen, H.~Fan, R.~Girshick, and K.~He.
\newblock Improved baselines with momentum contrastive learning.
\newblock {\em arXiv preprint arXiv:2003.04297}, 2020.

\bibitem{deng2009imagenet}
J.~Deng, W.~Dong, R.~Socher, L.-J. Li, K.~Li, and L.~Fei-Fei.
\newblock Image{N}et: {A} large-scale hierarchical image database.
\newblock In {\em CVPR}, 2009.

\bibitem{du2019EBM}
Y.~Du and I.~Mordatch.
\newblock Implicit generation and modeling with energy based models.
\newblock In {\em NeurIPS}, 2019.

\bibitem{fredrikson2015modelinversionattack}
M.~Fredrikson, S.~Jha, and T.~Ristenpart.
\newblock Model inversion attacks that exploit confidence information and basic
  countermeasures.
\newblock In {\em CCS}, 2015.

\bibitem{gao2018learning}
R.~Gao, Y.~Lu, J.~Zhou, S.-C. Zhu, and Y.~Nian~Wu.
\newblock Learning generative convnets via multi-grid modeling and sampling.
\newblock In {\em CVPR}, 2018.

\bibitem{geiping2020inverting}
J.~Geiping, H.~Bauermeister, H.~Dr{\"o}ge, and M.~Moeller.
\newblock Inverting gradients--{H}ow easy is it to break privacy in federated
  learning?
\newblock In {\em NeurIPS}, 2020.

\bibitem{zgeiping_github}
J.~Geiping, H.~Bauermeister, H.~Dr{\"o}ge, and M.~Moeller.
\newblock {Inverting Gradients Github Source Code}.
\newblock \url{https://github.com/JonasGeiping/invertinggradients}, 2020.
\newblock [Online; accessed 3-Nov-2020].

\bibitem{grathwohl2019your}
W.~Grathwohl, K.-C. Wang, J.-H. Jacobsen, D.~Duvenaud, M.~Norouzi, and
  K.~Swersky.
\newblock Your classifier is secretly an energy based model and you should
  treat it like one.
\newblock {\em ICLR}, 2020.

\bibitem{gulrajani2017improved}
I.~Gulrajani, F.~Ahmed, M.~Arjovsky, V.~Dumoulin, and A.~C. Courville.
\newblock Improved training of {W}asserstein {GAN}s.
\newblock In {\em NeurIPS}, 2017.

\bibitem{haroush2020knowledge}
M.~Haroush, I.~Hubara, E.~Hoffer, and D.~Soudry.
\newblock {T}he knowledge within: {M}ethods for data-free model compression.
\newblock In {\em CVPR}, 2020.

\bibitem{he2020momentum}
K.~He, H.~Fan, Y.~Wu, S.~Xie, and R.~Girshick.
\newblock Momentum contrast for unsupervised visual representation learning.
\newblock In {\em CVPR}, 2020.

\bibitem{he2016deep}
K.~He, X.~Zhang, S.~Ren, and J.~Sun.
\newblock Deep residual learning for image recognition.
\newblock In {\em CVPR}, 2016.

\bibitem{he2019model}
Z.~He, T.~Zhang, and R.~B. Lee.
\newblock Model inversion attacks against collaborative inference.
\newblock In {\em ACSAC}, 2019.

\bibitem{hitaj2017deep}
B.~Hitaj, G.~Ateniese, and F.~Perez-Cruz.
\newblock Deep models under the {GAN}: {I}nformation leakage from collaborative
  deep learning.
\newblock In {\em CCS}, 2017.

\bibitem{iandola2016firecaffe}
F.~N. Iandola, M.~W. Moskewicz, K.~Ashraf, and K.~Keutzer.
\newblock Firecaffe: {N}ear-linear acceleration of deep neural network training
  on compute clusters.
\newblock In {\em CVPR}, 2016.

\bibitem{karras2020analyzing}
T.~Karras, S.~Laine, M.~Aittala, J.~Hellsten, J.~Lehtinen, and T.~Aila.
\newblock Analyzing and improving the image quality of {S}tyle{GAN}.
\newblock In {\em CVPR}, 2020.

\bibitem{konevcny2016federated}
J.~Kone{\v{c}}n{\`y}, H.~B. McMahan, D.~Ramage, and P.~Richt{\'a}rik.
\newblock Federated optimization: {D}istributed machine learning for on-device
  intelligence.
\newblock {\em arXiv preprint arXiv:1610.02527}, 2016.

\bibitem{konevcny2016federated2}
J.~Kone{\v{c}}n{\`y}, H.~B. McMahan, F.~X. Yu, P.~Richt{\'a}rik, A.~T. Suresh,
  and D.~Bacon.
\newblock Federated learning: {S}trategies for improving communication
  efficiency.
\newblock {\em arXiv preprint arXiv:1610.05492}, 2016.

\bibitem{le2017privacy}
T.~Le, Y.~Aono, T.~Hayashi, et~al.
\newblock Privacy-preserving deep learning: {R}evisited and enhanced.
\newblock In {\em ICATIS}, pages 100--110, 2017.

\bibitem{lei2019inverting}
Q.~Lei, A.~Jalal, I.~S. Dhillon, and A.~G. Dimakis.
\newblock Inverting deep generative models, one layer at a time.
\newblock In {\em NeurIPS}, 2019.

\bibitem{li2014scaling}
M.~Li, D.~G. Andersen, J.~W. Park, A.~J. Smola, A.~Ahmed, V.~Josifovski,
  J.~Long, E.~J. Shekita, and B.-Y. Su.
\newblock Scaling distributed machine learning with the parameter server.
\newblock In {\em OSDI}, pages 583--598, 2014.

\bibitem{liu2018path}
S.~Liu, L.~Qi, H.~Qin, J.~Shi, and J.~Jia.
\newblock Path aggregation network for instance segmentation.
\newblock In {\em CVPR}, 2018.

\bibitem{mahendran2015understanding}
A.~Mahendran and A.~Vedaldi.
\newblock Understanding deep image representations by inverting them.
\newblock In {\em CVPR}, 2015.

\bibitem{mahendran2016visualizing}
A.~Mahendran and A.~Vedaldi.
\newblock Visualizing deep convolutional neural networks using natural
  pre-images.
\newblock {\em IJCV}, 2016.

\bibitem{mcmahan2017communication}
B.~McMahan, E.~Moore, D.~Ramage, S.~Hampson, and B.~A. y~Arcas.
\newblock Communication-efficient learning of deep networks from decentralized
  data.
\newblock In {\em AISTATS}.

\bibitem{melis2019exploiting}
L.~Melis, C.~Song, E.~De~Cristofaro, and V.~Shmatikov.
\newblock Exploiting unintended feature leakage in collaborative learning.
\newblock In {\em IEEE Symp. Security and Privacy (SP)}.

\bibitem{micaelli2019zero}
P.~Micaelli and A.~J. Storkey.
\newblock Zero-shot knowledge transfer via adversarial belief matching.
\newblock In {\em NeurIPS}, 2019.

\bibitem{micikevicius2017mixed}
P.~Micikevicius, S.~Narang, J.~Alben, G.~Diamos, E.~Elsen, D.~Garcia,
  B.~Ginsburg, M.~Houston, O.~Kuchaiev, G.~Venkatesh, and H.~Wu.
\newblock Mixed precision training.
\newblock In {\em ICLR}, 2018.

\bibitem{miyato2018spectral}
T.~Miyato, T.~Kataoka, M.~Koyama, and Y.~Yoshida.
\newblock Spectral normalization for generative adversarial networks.
\newblock In {\em ICLR}, 2018.

\bibitem{mordvintsev2015deepdream}
A.~Mordvintsev, C.~Olah, and M.~Tyka.
\newblock Inceptionism: {G}oing deeper into neural networks.
\newblock
  \url{https://research.googleblog.com/2015/06/inceptionism-going-deeper-into-neural.html},
  2015.

\bibitem{nguyen2017plug}
A.~Nguyen, J.~Clune, Y.~Bengio, A.~Dosovitskiy, and J.~Yosinski.
\newblock Plug \& play generative networks: {C}onditional iterative generation
  of images in latent space.
\newblock In {\em CVPR}, 2017.

\bibitem{nguyen2016synthesizing}
A.~Nguyen, A.~Dosovitskiy, J.~Yosinski, T.~Brox, and J.~Clune.
\newblock Synthesizing the preferred inputs for neurons in neural networks via
  deep generator networks.
\newblock In {\em NeurIPS}, 2016.

\bibitem{nguyen2015deep}
A.~Nguyen, J.~Yosinski, and J.~Clune.
\newblock Deep neural networks are easily fooled: {H}igh confidence predictions
  for unrecognizable images.
\newblock In {\em CVPR}, 2015.

\bibitem{samarakoon2018federated}
S.~Samarakoon, M.~Bennis, W.~Saad, and M.~Debbah.
\newblock Federated learning for ultra-reliable low-latency {V2V}
  communications.
\newblock In {\em GLOBECOM}, 2018.

\bibitem{santurkar2019image}
S.~Santurkar, A.~Ilyas, D.~Tsipras, L.~Engstrom, B.~Tran, and A.~Madry.
\newblock Image synthesis with a single (robust) classifier.
\newblock In {\em NeurIPS}, 2019.

\bibitem{shen2020ransac}
X.~Shen, F.~Darmon, A.~A. Efros, and M.~Aubry.
\newblock {RANSAC}-{F}low: {G}eneric two-stage image alignment.
\newblock In {\em ECCV}, 2020.

\bibitem{shokri2017membership}
R.~Shokri, M.~Stronati, C.~Song, and V.~Shmatikov.
\newblock Membership inference attacks against machine learning models.
\newblock In {\em IEEE Symp. Security and Privacy (SP)}.

\bibitem{wang2015regression}
Y.~Wang, C.~Si, and X.~Wu.
\newblock Regression model fitting under differential privacy and model
  inversion attack.
\newblock In {\em IJCAI}, 2015.

\bibitem{wang2019beyond}
Z.~Wang, M.~Song, Z.~Zhang, Y.~Song, Q.~Wang, and H.~Qi.
\newblock Beyond inferring class representatives: User-level privacy leakage
  from federated learning.
\newblock In {\em INFOCOM}, 2019.

\bibitem{yang2019adversarial}
Z.~Yang, E.-C. Chang, and Z.~Liang.
\newblock Adversarial neural network inversion via auxiliary knowledge
  alignment.
\newblock {\em arXiv preprint arXiv:1902.08552}, 2019.

\bibitem{yin2020dreaming}
H.~Yin, P.~Molchanov, J.~M. Alvarez, Z.~Li, A.~Mallya, D.~Hoiem, N.~K. Jha, and
  J.~Kautz.
\newblock Dreaming to distill: Data-free knowledge transfer via
  {D}eep{I}nversion.
\newblock In {\em CVPR}.

\bibitem{zhang2018context}
H.~Zhang, K.~Dana, J.~Shi, Z.~Zhang, X.~Wang, A.~Tyagi, and A.~Agrawal.
\newblock Context encoding for semantic segmentation.
\newblock In {\em CVPR}, 2018.

\bibitem{zhang2018self}
H.~Zhang, I.~Goodfellow, D.~Metaxas, and A.~Odena.
\newblock Self-attention generative adversarial networks.
\newblock In {\em ICML}, 2019.

\bibitem{zhang2018perceptual}
R.~Zhang, P.~Isola, A.~A. Efros, E.~Shechtman, and O.~Wang.
\newblock The unreasonable effectiveness of deep features as a perceptual
  metric.
\newblock In {\em CVPR}, 2018.

\bibitem{zhang2020secret}
Y.~Zhang, R.~Jia, H.~Pei, W.~Wang, B.~Li, and D.~Song.
\newblock The secret revealer: {G}enerative model-inversion attacks against
  deep neural networks.
\newblock In {\em CVPR}, 2020.

\bibitem{zhao2020idlg}
B.~Zhao, K.~R. Mopuri, and H.~Bilen.
\newblock i{DLG}: {I}mproved deep leakage from gradients.
\newblock {\em arXiv preprint arXiv:2001.02610}, 2020.

\bibitem{zhao2017pyramid}
H.~Zhao, J.~Shi, X.~Qi, X.~Wang, and J.~Jia.
\newblock Pyramid scene parsing network.
\newblock In {\em CVPR}, 2017.

\bibitem{zhu2019deep}
L.~Zhu, Z.~Liu, and S.~Han.
\newblock Deep leakage from gradients.
\newblock In {\em NeurIPS}, 2019.

\bibitem{zzhu_github}
L.~Zhu, Z.~Liu, and S.~Han.
\newblock {Deep Leakage From Gradients Github Source Code}.
\newblock \url{https://github.com/mit-han-lab/dlg}, 2019.
\newblock [Online; accessed 3-Nov-2020].

\end{thebibliography}
}

\appendix
\noindent\begin{figure*}[!t]

\begingroup
\renewcommand*{\arraystretch}{0.3}
\begin{tabular}{l}
\large{\textbf{Appendix A - More Examples}}
\vspace{3mm}
\end{tabular}
\endgroup

\centering
\resizebox{.92\linewidth}{!}{
\begingroup
\renewcommand*{\arraystretch}{0.3}
\begin{tabular}{ccc}
\includegraphics[width=0.33\linewidth,clip,trim=5px 0 0 4px]{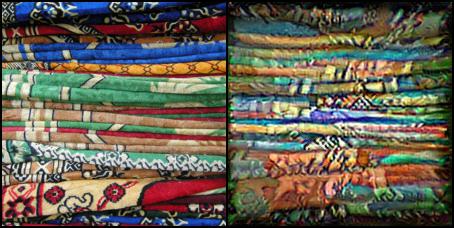} &
\includegraphics[width=0.33\linewidth,clip,trim=5px 0 0 4px]{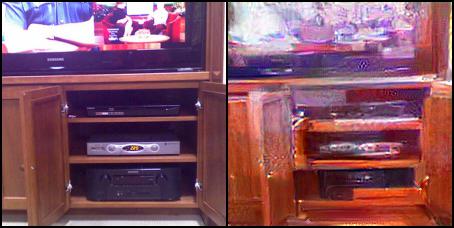} &
\includegraphics[width=0.33\linewidth,clip,trim=5px 0 0 4px]{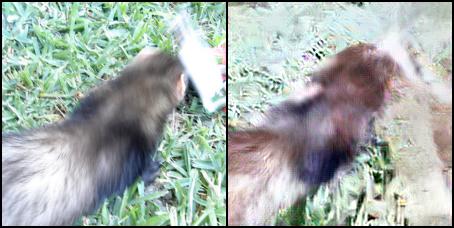}
\\
\includegraphics[width=0.33\linewidth,clip,trim=5px 0 0 4px]{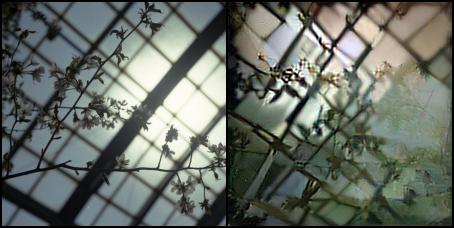} &
\includegraphics[width=0.33\linewidth,clip,trim=5px 0 0 4px]{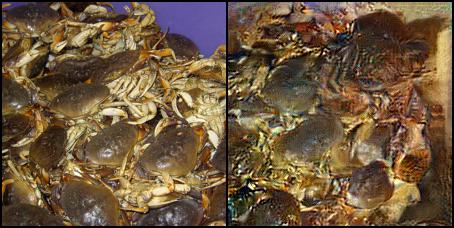} &
\includegraphics[width=0.33\linewidth,clip,trim=5px 0 0 4px]{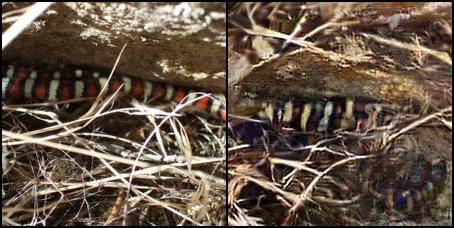}
\\
\multicolumn{3}{c}{batch size $4$}
\vspace{1mm}
\end{tabular}
\endgroup
}

\resizebox{.92\linewidth}{!}{
\begingroup
\renewcommand*{\arraystretch}{0.3}
\begin{tabular}{ccc}
\includegraphics[width=0.33\linewidth,clip,trim=5px 0 0 4px]{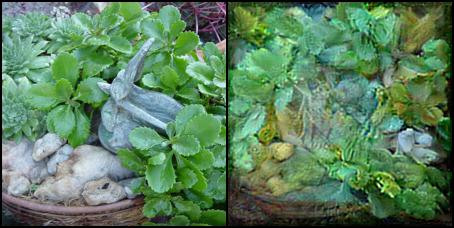} &
\includegraphics[width=0.33\linewidth,clip,trim=5px 0 0 4px]{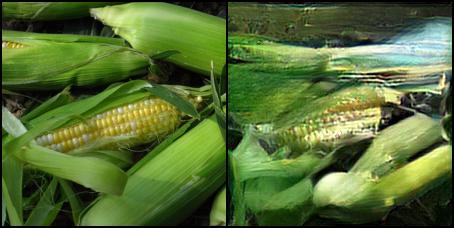} &
\includegraphics[width=0.33\linewidth,clip,trim=5px 0 0 4px]{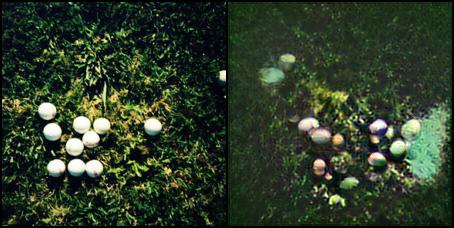}
\\
\includegraphics[width=0.33\linewidth,clip,trim=5px 0 0 4px]{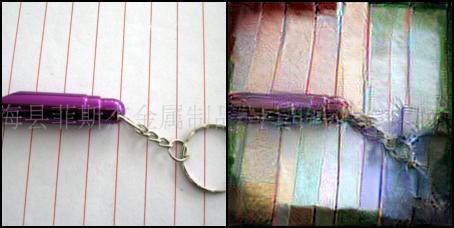} &
\includegraphics[width=0.33\linewidth,clip,trim=5px 0 0 4px]{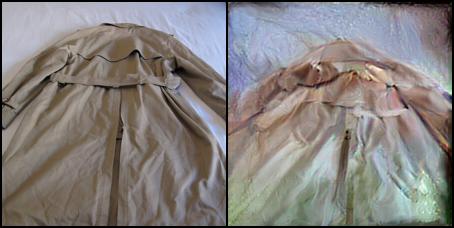} &
\includegraphics[width=0.33\linewidth,clip,trim=5px 0 0 4px]{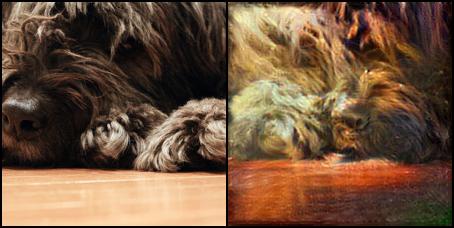}
\\
\multicolumn{3}{c}{batch size $8$}
\vspace{1mm}
\end{tabular}
\endgroup
}

\resizebox{.92\linewidth}{!}{
\begingroup
\renewcommand*{\arraystretch}{0.3}
\begin{tabular}{ccc}
\includegraphics[width=0.33\linewidth,clip,trim=5px 0 0 4px]{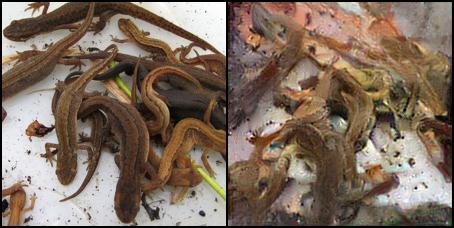} &
\includegraphics[width=0.33\linewidth,clip,trim=5px 0 0 4px]{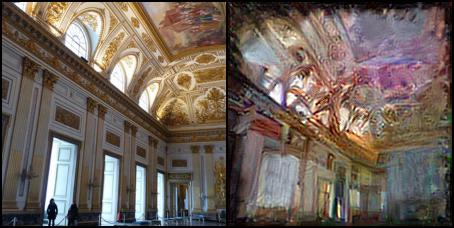}  &
\includegraphics[width=0.33\linewidth,clip,trim=5px 0 0 4px]{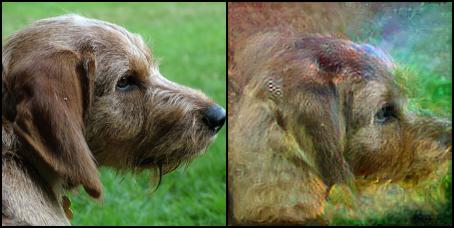}  
\\
\includegraphics[width=0.33\linewidth,clip,trim=5px 0 0 4px]{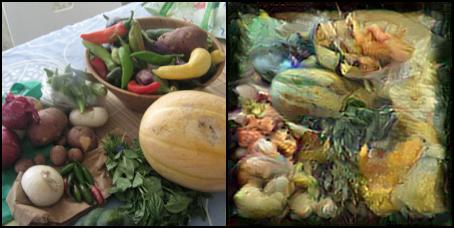}  &
\includegraphics[width=0.33\linewidth,clip,trim=5px 0 0 4px]{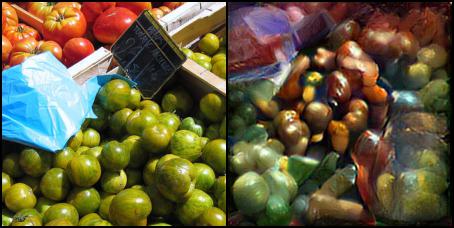}  &
\includegraphics[width=0.33\linewidth,clip,trim=5px 0 0 4px]{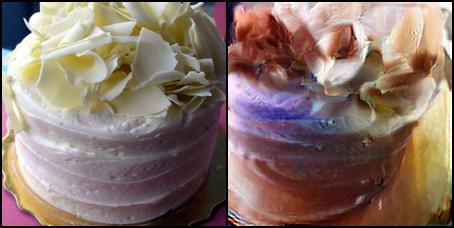}  
\\
\multicolumn{3}{c}{batch size $16$}
\vspace{1mm}
\end{tabular}
\endgroup
}
    
\resizebox{.92\linewidth}{!}{
\begingroup
\renewcommand*{\arraystretch}{0.3}
\begin{tabular}{ccc}
\includegraphics[width=0.33\linewidth,clip,trim=5px 0 0 4px]{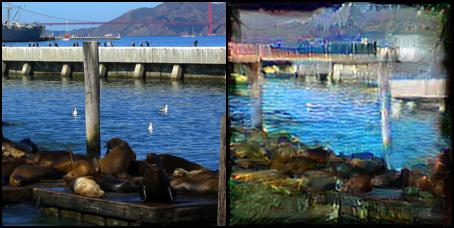} &
\includegraphics[width=0.33\linewidth,clip,trim=5px 0 0 4px]{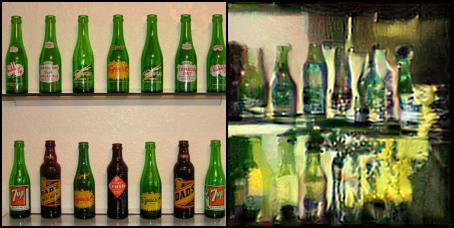} &
\includegraphics[width=0.33\linewidth,clip,trim=5px 0 0 4px]{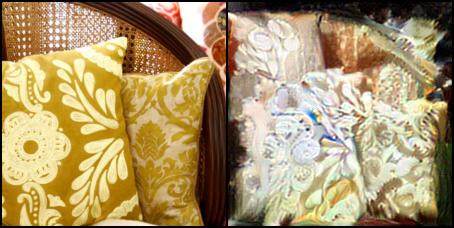} 
\\
\includegraphics[width=0.33\linewidth,clip,trim=5px 0 0 4px]{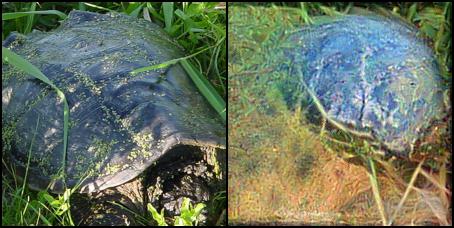} &
\includegraphics[width=0.33\linewidth,clip,trim=5px 0 0 4px]{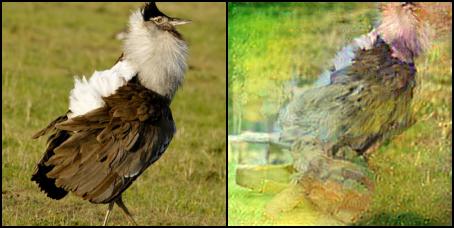} &
\includegraphics[width=0.33\linewidth,clip,trim=5px 0 0 4px]{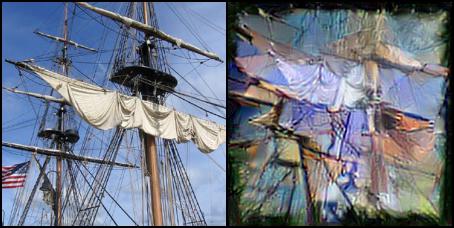} 
\\
\multicolumn{3}{c}{batch size $32$}
\vspace{1mm}
\end{tabular}
\endgroup
}
\caption{Additional examples of information leakage when inverting ResNet-50 gradients on the ImageNet validation set. Each block containing a pair of (left) original sample and its (right) reconstruction by GradInversion. }
\label{fig:more_and_more}
\end{figure*}
\begin{figure*}[!t]

\begingroup
\renewcommand*{\arraystretch}{0.3}
\begin{tabular}{l}
\large{\textbf{Appendix B - Ablation Studies Images}}
\vspace{1cm}
\end{tabular}
\endgroup

\centering
\resizebox{\linewidth}{!}{
\begingroup
\renewcommand*{\arraystretch}{0.3}
\begin{tabular}{c}
\includegraphics[width=.95\linewidth]{figures/ablation_images/original.jpg} \\
 \textbf{Original batch - ground truth} 
\vspace{2mm}\\ 
\midrule
\vspace{2mm}
\end{tabular}
\endgroup
}

\resizebox{\linewidth}{!}{
\begingroup
\renewcommand*{\arraystretch}{0.3}
\begin{tabular}{c}
\includegraphics[width=.95\linewidth]{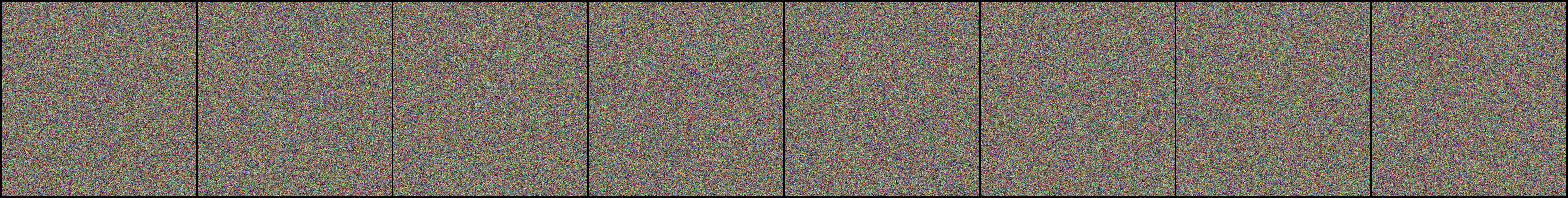} \\
 Noise $\mathcal{N}(0, \mathcal{I})$ - starting point 
\vspace{2mm}
\end{tabular}
\endgroup
}

\resizebox{\linewidth}{!}{
\begingroup
\renewcommand*{\arraystretch}{0.3}
\begin{tabular}{c}
\includegraphics[width=.95\linewidth]{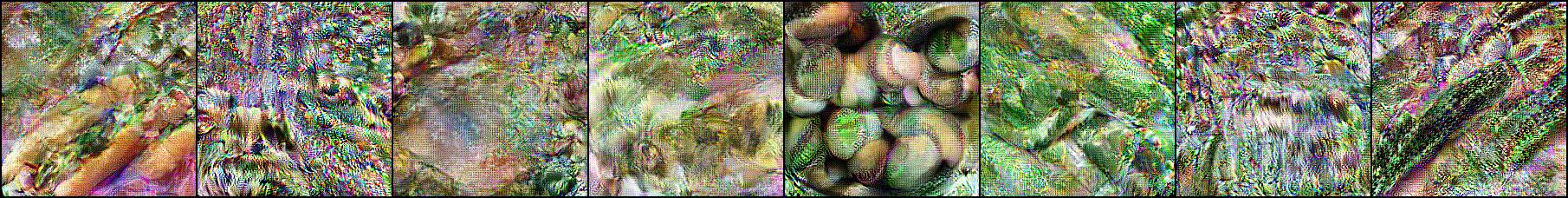} \\
$\mathcal{L_\text{grad}}$\\
\vspace{1mm}
\end{tabular}
\endgroup
}

\resizebox{\linewidth}{!}{
\begingroup
\renewcommand*{\arraystretch}{0.3}
\begin{tabular}{c}
\includegraphics[width=.95\linewidth]{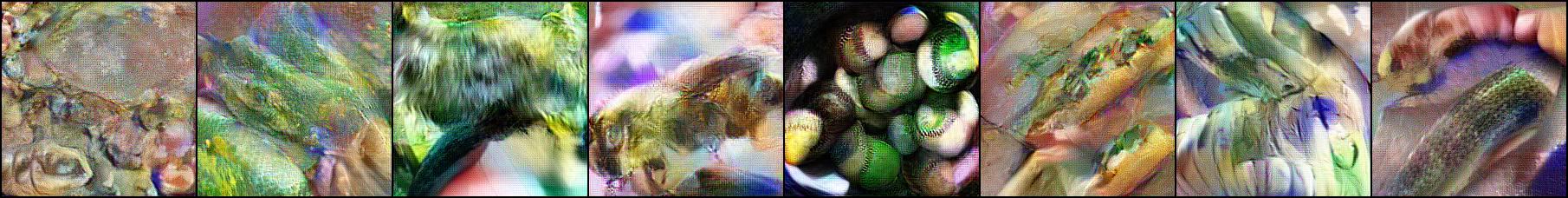} \\
+ $\mathcal{R_\text{fidelity}}$ \\
\vspace{1mm}
\end{tabular}
\endgroup
}

\resizebox{\linewidth}{!}{
\begingroup
\renewcommand*{\arraystretch}{0.3}
\begin{tabular}{c}
\includegraphics[width=.95\linewidth]{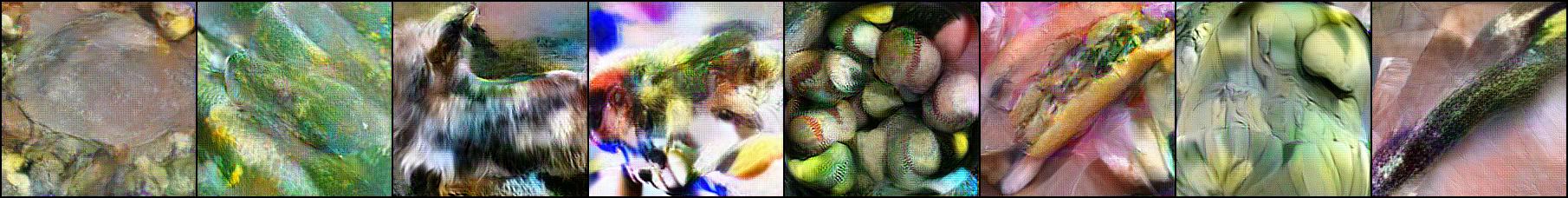} \\
+ $\mathcal{R_\text{group.lazy}}$ \\
\vspace{1mm}
\end{tabular}
\endgroup
}

\resizebox{\linewidth}{!}{
\begingroup
\renewcommand*{\arraystretch}{0.3}
\begin{tabular}{c}
\includegraphics[width=.95\linewidth]{figures/ablation_images/4gpu_reg2.jpg} \\
+ $\mathcal{R_\text{group.reg}}$ \textbf{(our final reconstruction)} \\
\vspace{3mm}
\end{tabular}
\endgroup
}
\caption{Detailed visual comparison when adding individual loss terms to GradInversion (supplementary for Table~\ref{tb:loss_ablation_table} of main paper).
Starting from noise, the gradient loss produces noisy outputs which begin to show glimpses of what the original image contains. The fidelity loss encourages the optimization to produce more realistic outputs. Using multiple random seeds and regularizing the inputs using even the simple lazy scheme of conforming to the mean image improves the image quality. Our group-based regularization that uses image registration produces the best-looking outputs.
}
\label{fig:ablation_imgs}
\end{figure*}

\newpage
\clearpage
\section*{Appendix C - Additional Details \& Analysis}

\paragraph{$\mathcal{L}_{grad}$ cost function.} For the task of gradient matching, we study how the loss function affects optimization in Table~\ref{tb:loss_ablation_table}. 
We find $\ell_2$ loss outperforms cosine similarity~\cite{geiping2020inverting} for gradient matching. To this end, we compare the $\ell_2$ distance, percentage of gradient signs that matched, and the cosine similarity between the final optimized gradient and the ground truth gradient. It can be observed that the $\ell_2$ loss results in stronger convergence for this task. 
\begin{table}[h]
\centering
\resizebox{.8\linewidth}{!}{
\begin{tabular}{lccc}
\toprule
\multirow{2}{*}{$\mathbf{\mathcal{L}}_\text{grad}(\cdot)$} & \multicolumn{3}{c}{\textbf{$\nabla_{\mathbf{W}}\mathcal{L}(\mathbf{\hat{x}^*}, \mathbf{\hat{y}^*})$ vs. $\nabla_{\mathbf{W}}\mathcal{L}(\mathbf{x^*}, \mathbf{y^*})$}}\\
\cmidrule{2-4}
 & $\ell_2$ dist. $\downarrow$ & sign (\%) $\uparrow$  & cos. dist. $\downarrow$ \\
\midrule
cosine~\cite{geiping2020inverting}      & 5.965 & 79.0 & 0.139 \\
$\ell_2$ \textbf{(ours)}    & \textbf{3.835} & \textbf{80.9} & \textbf{0.110} \\
\bottomrule
\end{tabular}}
\label{tab:vs_cos}
\caption{Comparing the efficacy for cosine and $\ell_2$ distance for gradient matching. }
\label{tb:loss_ablation_table}
\end{table}





\begin{figure}[t]
\centering
\resizebox{\linewidth}{!}{
\begingroup
\renewcommand*{\arraystretch}{0.3}
\begin{tabular}{cc}
\includegraphics[width=0.5\linewidth,clip,trim=5px 0 0 4px]{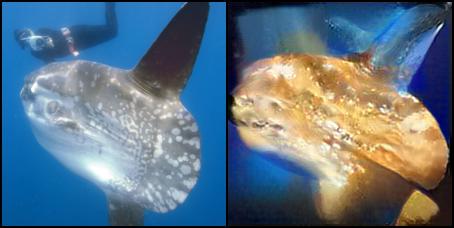} &
\includegraphics[width=0.5\linewidth,clip,trim=5px 0 0 4px]{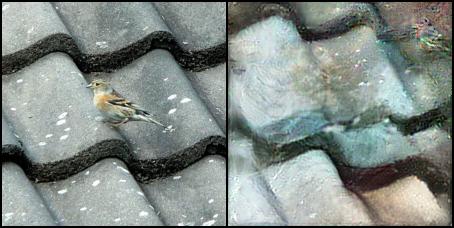} \\
\multicolumn{2}{c}{(a) Vanishing objects.}
\vspace{2mm}
\end{tabular}
\endgroup
}

\resizebox{\linewidth}{!}{
\begingroup
\renewcommand*{\arraystretch}{0.3}
\begin{tabular}{cc}
\includegraphics[width=0.5\linewidth,clip,trim=5px 0 0 4px]{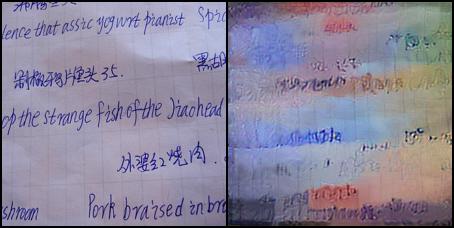} &
\includegraphics[width=0.5\linewidth,clip,trim=5px 0 0 4px]{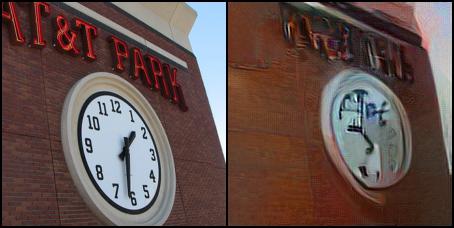} \\
\multicolumn{2}{c}{(b) Blurred text \& digits.}
\vspace{1mm}
\end{tabular}
\endgroup
}

\resizebox{\linewidth}{!}{
\begingroup
\renewcommand*{\arraystretch}{0.3}
\begin{tabular}{cc}
\includegraphics[width=0.5\linewidth,clip,trim=5px 0 0 4px]{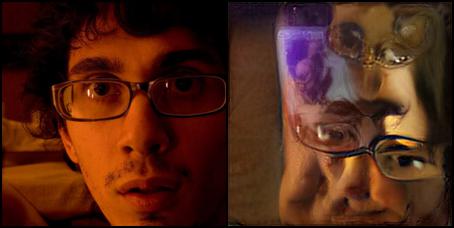} &
\includegraphics[width=0.5\linewidth,clip,trim=5px 0 0 4px]{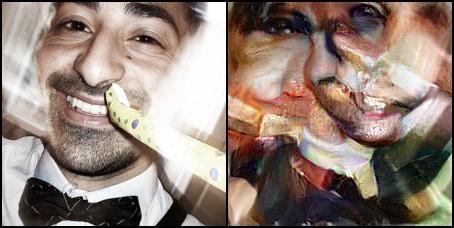} \\
\multicolumn{2}{c}{(c) Human faces.}\\
\vspace{1mm}
\end{tabular}
\endgroup
}


\caption{Insights and observations of GradInversion given ResNet-50 gradients on ImageNet. Each block containing a pair of (left) original sample and its (right) reconstruction by GradInversion. Samples from inversion results at batch sizes $4$ and $8$.}
\label{fig:discussions}
\end{figure}

\paragraph{Insights \& open challenges.} We next discuss several of our observations when performing GradInversion for the ResNet-50 network (MOCO V2) on the ImageNet1K dataset. 
We would like to note that these observations hold for the chosen network and optimization settings used, and may not be general in scope.
We hope that sharing our experiences would help provide insights for future work. 
\begin{itemize}
    \item \textbf{Vanishing objects.} Images recovered from gradient inversion occasionally omit details of original images, as shown in Fig.~\ref{fig:discussions} (a) where the diver and the bird disappear post inversion. The observation is in line with the missing details phenomena during the latent code projection as in StyleGAN2 by Karras \etal~\cite{karras2020analyzing}.
    
    \item \textbf{Texts \& digits.} GradInversion unveils existence of texts and digits, while their exact details remain blurry. See several quick examples in Fig.~\ref{fig:discussions} (b).
    
    \item \textbf{Human faces.} Recovery remains harder for samples and distributions in ImageNet that involve human faces (also deemed challenging by BigGAN~\cite{brock2018biggan}). Even though detailed features and patches can be reversed, \eg, mouths, eyes, and noses, \etc, they are not correctly arranged spatially, as shown in Fig.~\ref{fig:discussions} (c). We conjecture that this is a result of (i) the under-representation of such distributions in ImageNet1K and  (ii) such features being ignored by the network for the classification task. Different observations may hold for other datasets where tasks enforce networks to focus on spatial alignments of facial features, \eg, towards facial recognition.
\end{itemize}

\end{document}